\documentclass[final,5p,times]{elsarticle}

\usepackage{amssymb}

\usepackage{lineno}
\usepackage{vruler}

\usepackage{amsmath}
\usepackage{xspace}
\usepackage[tight,footnotesize]{subfigure}
\usepackage{multirow}
\usepackage{algorithm}
\usepackage{algpseudocode}
\usepackage{bm}
\usepackage{upgreek}
\usepackage[table]{xcolor}
\usepackage{graphicx}
\usepackage{amsmath,amssymb}
\usepackage[nolist,nohyperlinks]{acronym}
\usepackage{ifthen}
\usepackage{xspace}
\usepackage{rotating}
\usepackage{todonotes}
\usepackage{multirow}
\usepackage{epstopdf}
\usepackage{amssymb}
\usepackage{xcolor}
\usepackage{microtype}
\usepackage{afterpage}

\mathchardef\mhyphen="2D

\journal{CVIU: }

\begin{document}

\begin{frontmatter}


\title{Image and Video Mining through Online Learning}


\author[surrey]{Andrew Gilbert\corref{cor1}}
\ead{a.gilbert@surrey.ac.uk}
\author[surrey]{Richard Bowden}
\ead{R.Bowden@surrey.ac.uk}
\address[surrey]{CVSSP, University of Surrey, GU2\,7XH Guildford, United Kingdom}
\cortext[cor1]{Corresponding author. Telephone: +44 1483 682260.}


\begin{abstract}
Within the field of image and video recognition, the traditional approach is a dataset split into fixed training and test partitions. However, the labelling of the training set is time-consuming, especially as datasets grow in size and complexity. Furthermore, this approach is not applicable to the home user, who wants to intuitively group their media without tirelessly labelling the content. Consequently, we propose a solution similar in nature to an active learning paradigm, where a small subset of media is labelled as semantically belonging to the same class, and machine learning is then used to “pull” this and other related content together in the feature space. Our interactive approach is able to iteratively cluster classes of images and video. We reformulate it in an online learning framework and demonstrate competitive performance to batch learning approaches using only a fraction of the labelled data. Our approach is based around the concept of an image signature which, unlike a standard bag of words model, can express co-occurrence statistics as well as symbol frequency. We efficiently compute metric distances between signatures despite their inherent high dimensionality and provide discriminative feature selection, to allow common and distinctive elements to be identified from a small set of user labelled examples. These elements are then accentuated in the image signature to increase similarity between examples and “pull” correct classes together. By repeating this process in an online learning framework, the accuracy of similarity increases dramatically despite labelling only a few training examples. To demonstrate that the approach is agnostic to media type and features used, we evaluate on three image datasets (15 scene, Caltech101 and FG-NET), a mixed text and image dataset (ImageTag), a dataset used in active learning (Iris) and on three action recognition datasets (UCF11, KTH and Hollywood2). On the UCF11 video dataset, the accuracy is 86.7\% despite using only 90 labelled examples from a dataset of over 1200 videos, instead of the standard 1122 training videos. The approach is both scalable and efficient, with a single iteration over the full UCF11 dataset of around 1200 videos taking approximately 1 minute on a standard desktop machine.

\end{abstract}
\end{frontmatter}

\section{Introduction}
Fuelled by the prevalence of cameras on mobile devices and social networking sites such as Facebook, Twitter and YouTube, digital content is ever increasing. This produces a demand for automatic approaches
to clustering media into meaningful semantic groups to facilitate browsing and search. This use case is incompatible with traditional supervised training methods, as labelling the data is the limiting factor. Therefore, we propose an approach that allows the user to find natural groups of similar content based on a small handful of ”seed” examples. Combining these seed examples with an automatic data mining approach that extracts rules that can generalise and further cluster the remaining unseen media.

There have been many approaches that are successful
in the classification of images and videos ~\cite{MarszalekCVPR09, GilbertICCV09,Schuldt04,WangBMVC09,HanICCV09}. However, these require significant amounts of supervised training data, which is increasingly infeasible to provide. There are ”single shot” approaches
that take a limited training set~\cite{ShechtmanPAMI07,NingCir08}. However, they can be sensitive to noise in the training data, and are difficult to generalise to larger datasets.

Conversely, we use an online learning approach capable of incrementally clustering similar material from the manual identification of a few correct and incorrect examples. These examples are then used to learn rules that can be applied to clustering a larger corpus of material. The approach is demonstrated on three pure image datasets
(15 Scene\cite{LazebnikCVPR06}, Caltech101\cite{FeiFeiCVPR04}, FG-NET~\cite{panis2015overviewFG-NET}), on a combined text and
image dataset (ImageTag~\cite{GilbertACCV12}), a dataset used in active
learning (Iris~\cite{Lichman2013}) and on three state-of-the-art video
action recognition datasets (UCF11\cite{LiuCVPR09}, KTH\cite{Schuldt04}, and Hollywood2\cite{MarszalekCVPR09}).

To provide both scalability and incremental learning, the approach needs to remain efficient as datasets become larger. Therefore, we efficiently compute both distances between high dimensional representations and dynamically augment the representation with new compound elements to form an image signature. We demonstrate the approach is independent of the underlying features. The similarity measure employed in this paper extends the original min-Hash algorithm that was designed to identify the similarity between text in documents~\cite{BroderSEQS98} by efficiently computing the distances between high-dimensional sets. Chum~\cite{ChumCIVR07} demonstrated the ability of min-Hash to efficiently identify near duplicate images within datasets. Min-Hash is ideally suited to large high dimensional representations, as the computational costs are not proportional to the size of the input representation. This makes it especially suited to complex image or video descriptors which are typically of high dimensionality. Chum~\cite{ChumBMVC08} later extended this work to approximate the histogram intersection of images.

Another data mining tool employed in this work
is association rule mining (known as APriori~\cite{Agrawal94})).
This was originally designed to identify co-occurring elements in large text files. It was first employed in the image domain by Quack~\cite{Quack07}. They used association rule mining in supervised object recognition to find spatially grouped SIFT descriptors.

In the temporal domain, Gilbert~\cite{GilbertICCV09} demonstrated the use of APriori in Action recognition. They argued that many other action recognition approaches ~\cite{Laptev03,LaptevCVPR08,Dollar05},  use features engineered to fire sparsely, to ensure that the overall problem is tractable. However, they suggested that this can sacrifice recognition accuracy as it cannot be assumed that the optimum features for class discrimination are obtained from this approach. In contrast, an over complete set of Harris corners~\cite{Harris88} are grouped spatially and temporally, mining is then used to identify
feature combinations to classify video sequences. While this demonstrated the power of APriori in
activity recognition, the training was still performed
with comprehensive supervised training sets.

\section{Related Works}
There is a number of related works that aim to reduce the labeleing of the training data. An online incremental algorithm (such as Law~\cite{law2004simultaneous}) can reduce the training examples and time required, we propose to include both correct and incorrect instances in a human led iterative process to select fewer but more relevant training examples. As with any approach that clusters or correlates images and video, the choice of the representation and similarity measure is critical, as they can affect both the size of
the database and the search time. We introduce the image signature as an efficient representation irrespective of the type of the input sample: image or video or the feature descriptor applied. Then, using APriori, the
distinctive and discriminative elements of these selected examples are identified and accentuated across the dataset by dynamically augmenting the representation with new compound elements. This increases the set overlap of correct image signatures while also improving the dissimilarity of incorrectly classified examples thereby increasing the overall accuracy of matching.
As the image signature increases in dimensionality,
min-Hash provides a scalable approach to computing similarity between data items. This iterative procedure
can be seen as a form of online learning with
similarities to approaches in both active and metric
learning.

Tong proposed active learning for the purpose of image retrieval~\cite{ Tong01supportvector}. Active learning is a particular case of semi-supervised machine learning where the learning algorithm interactively queries the user to obtain the desired outputs for new data points. Since the learner can identify examples of great confusion or variation to focus on, the number of examples to label for a concept can often be much lower than the number required in batch. This is a key aspect of our approach, in classical active learning, the algorithm chooses the
data points to be labelled based on some automated criteria. Our approach uses the notion of similarity and allows the user to select obvious outliers that should be labelled. Similarity helps the user prioritise annotation, and the feature representation is manipulated to satisfy these constraints. This changes the topology of the distance space and is therefore also related to Metric Learning. Metric learning is the task of learning a distance function over a dataset usually pairwise metric distances between samples.

There have recent developments involving users in hybrid active learning approaches~\cite{lughofer2012hybrid,Weigl_2016_AL_IKNOW,Weigl_2016_MVA}. \cite{lughofer2012hybrid} employs sample selection in the first phase based purely on unsupervised criteria. Then in the second phase, the task is to update the pre-trained classifiers with the most relevant samples. We propose a similar ideology however allow the user to select the “relevant samples” via a Multi-Dimensional Scaling (MDS) visualisation and unsupervised clustering of the distance between all data samples together with the novel approach identification of the discriminative features. While \cite{Weigl_2016_AL_IKNOW} is similar to this work through allowing the user to select the most relevant samples based on a visualization map showing the sample/class distributions. However we propose a more generic feature type to ensure multiple data models can be incorporated in this single method. \cite{Weigl_2016_MVA} performs on-line image classification tasks, in this case for event type classification, presenting the user "questionable" events for the user to examine instead of the whole dataset. Although the speed and ease of visualisation and the feature learning within this approach allows the full datasets to be presented to the user at iteration to ensure they don't get stuck in a local minima in the dataset.


\subsection{Paper Overview}
In this manuscript, we build upon our previous work
in ~\cite{GilbertICCV11,GilbertBMVC11} which introduced the online learning framework and was combined with a hand gesture estimation controller~\cite{KrejovCGA14}. This manuscript provides a mature and a detailed description of the approach. We have reformulated the learning framework and provide an extensive formalisation of the method to allow for repeatability. Regarding analysis, additional features have
been added and evaluated on seven different datasets, which include a broad range of various modalities (i.e. image, video and combined image/text-tag) using multiple user runs. We also provide analysis regarding cluster purity and evaluation of the computational cost
of the approach, showing that the online learning framework can compete favourably with the state of the art supervised learning approaches using only a fraction of the data.

Section 2 introduces the image signature and extends the min-Hash algorithm for video similarity in section 3. An image signature is a symbolised vector suitable for use by frequency based mining algorithms. The process of symbolisation takes a fixed dimensionality
vector, such as a histogram, and converts it into a variable length set of discrete symbols. Each symbol represents a dimension in the original vector, the number of times each symbol appears relates to the magnitude of that dimension. The learning framework is described with clustering and visualisation discussed in Section 4. Section 5 illustrates how frequent itemset mining can be modified to identify
discriminative or common elements of the signatures, that are then accentuated (section 6) to change the topology of the feature space. Extensive results are then provided on seven image and video datasets in section~\ref{sec:Results}.
\section{Overview}
\label{InputSig}

Previous approaches to the classification of video and
images, often use local feature point detectors and
descriptors to provide a compact representation~\cite{Sift,Dollar05,DalahCVPR05,Chatfield14_CNN}. Desirable properties are invariance to illumination
and geometric transformations. The descriptors
are often quantized, by clustering into a smaller set
of visual words, otherwise known as a code book or
”bag of words” (BoW)~\cite{FeiFeiCVPR05,SivicICCV05}. However, rather than
using a static BoWs histogram, we propose a dynamic
variant called an image signature.

The image signature has similarities to a classical
BoW in that it uses the frequency histogram of a set of discrete elements; it differs by being able to increase in size, to accentuate elements or features that are found to discriminate between classes. The signature is based on the response of any feature classifier. Initially,
we describe a signature based on a BoW model but
later we demonstrate its application to other classifier responses for both images, video and text.

An image signature is constructed for each data
item as the frequency of features extracted from the
data. This unique signature provides a compact, discrete
representation of the input sample. The initial
signature is effectively a standard BoW. However, a
new set of symbols is appended to the histogram
at each iteration of learning. These new symbols
represent compound combinations of previously co-occurring
elements. Compound elements are identified through the APriori data mining stage, to provide
additional rules that will bring examples of media
from the same class, closer regarding their similarity.

Figure 1 gives an overview of the approach. For each item, extracted features are converted into image signatures (Sec. 3.2) to form the initial signature database. From this database, pairwise distances are computed between all signatures (Sec. 3.1) and projected consistently to a visualisation space via  multidimensional scaling (MDS) (Sec. 4). The MDS presents the data to the user as a two or three-dimensional projection into Euclidean space with the similarity represented by proximity and groups highlighted via agglomerative clustering. The user then selects a limited number of items that should form either the same or different
classes and features within the signatures that satisfy these constraints are identified automatically (Sec. 5). All signatures in the database are then adjusted in light of these new rules (Sec. 6). This has the effect of pulling the signatures from the correct examples closer together. This process is then repeated, allowing a user to cluster their data iteratively by concentrating
on areas of apparent confusion.

%
%

\begin{figure}[htbp]
\centering
\includegraphics[width = 1 \columnwidth]{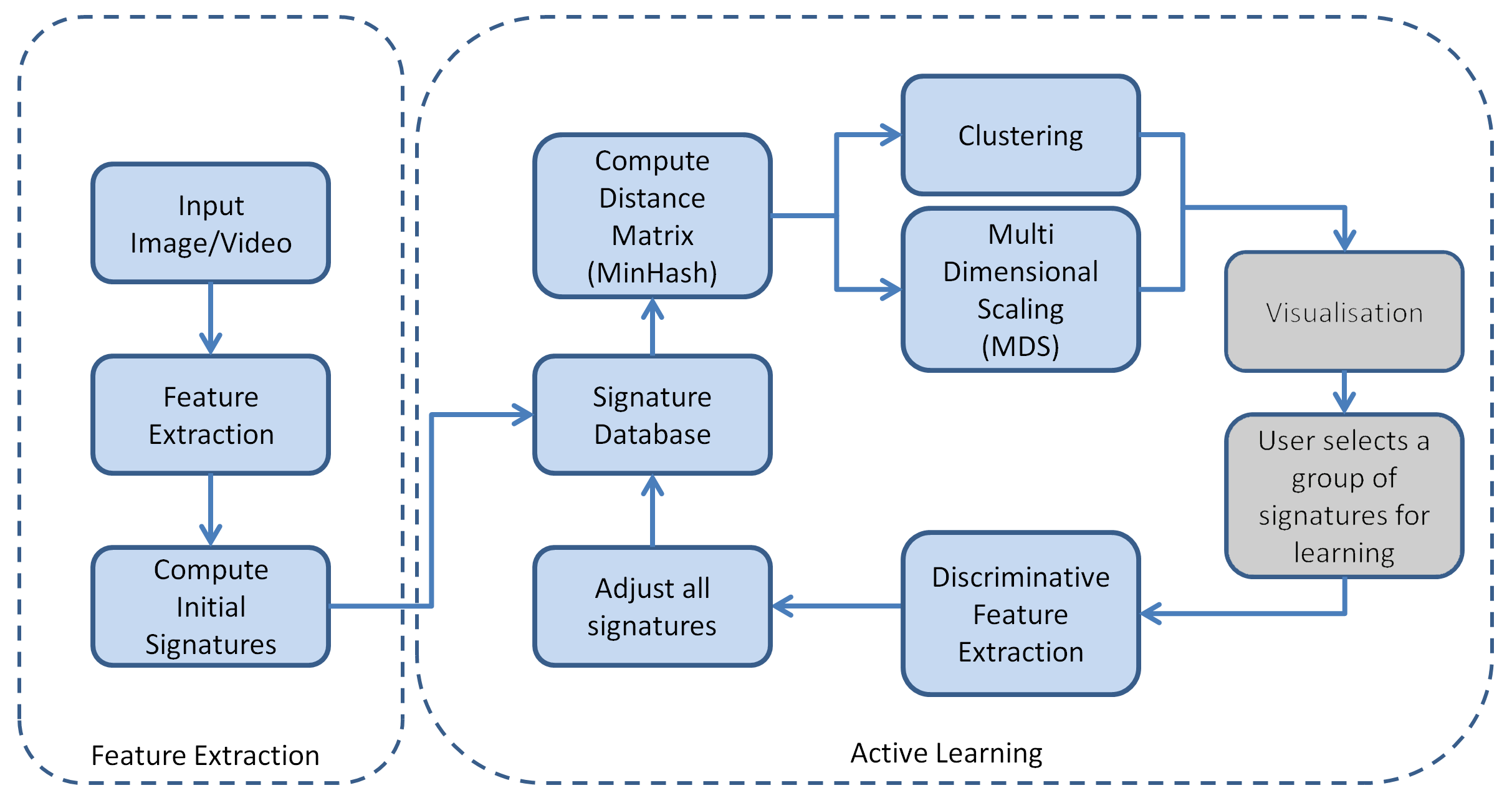}
\caption{ An overview of the Learning Framework - blue
depicts automated steps, gray those that involve the user.} \label{fig:Overview}
\end{figure}
\section{Similarity of signatures}
\label{SimSig}
The approach requires that the pair-wise similarity between the image signatures are computed efficiently, as learning needs the similarity of all signatures to be calculated at each iteration. The image signatures are large one-dimensional containers and to calculate the similarity efficiently is a challenging proposal; therefore we adapt the data mining tool, min-Hash as this can correlate long sets of symbols efficiently. Min-Hash was originally
developed for near-duplicate detection of large text
passages~\cite{BroderSEQS98} and more recently adopted for
the near duplicate detection of large image sets~\cite{ChumBMVC08}.
We extend this work efficiently to calculate the pair-wise similarity of image signatures. It estimates the set
overlap of pairs of sets, through randomised hashes
taken from the overall vocabulary of features. Min-Hash has the valuable property that the computation is proportional to the number of sets or samples rather
than the complexity of the vocabulary. As such, it is
ideally suited for use with image signatures which can
be of high and increasing dimensionality.

\subsection{The min-Hash algorithm}
\label{minHashAlgorithm}
\begin{figure}[htbp]
\centering
\includegraphics[width = 1 \columnwidth]{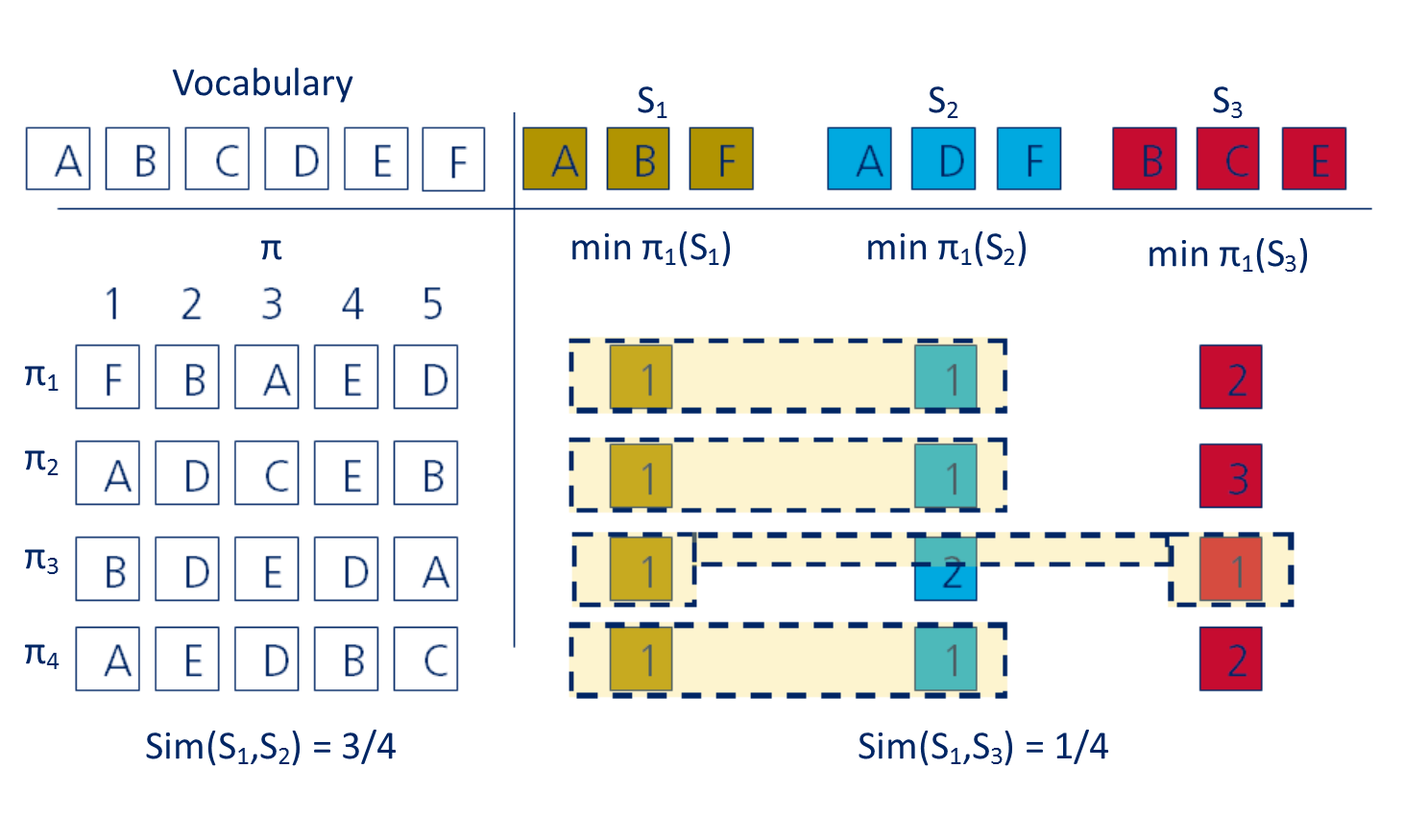}
\caption{An overview of the proposed approach} \label{fig:minhash}
\end{figure}

The distance similarity measure between two input samples is computed as the similarity of signature $\textbf{S}_{1}$ and $\textbf{S}_{2}$, the ratio of the number of features or elements in the intersection, over the union of the two signatures.
\begin{equation}
    sim(\textbf{S}_{1},\textbf{S}_{2}) = \frac{|\textbf{S}_{1}\cap \textbf{S}_{2}|}{|\textbf{S}_{1}\cup \textbf{S}_{2}|}
\end{equation}
Min-Hash is able to estimate $sim(\textbf{S}_{1},\textbf{S}_{2})$ without performing an exhaustive naive element by element comparison of $\textbf{S}_{1}$ and $\textbf{S}_{2}$. Instead, a set of random hash permutations $\textbf{N}=\{\pi_1, ... ,\pi_{|\textbf{N}|} \}$, of the vocabulary of elements, $\boldsymbol{\nu}$, are created. Each element in each random hash is sequentially examined in turn to see if it occurs within each image signature. If the element is found in the image signature, the index of the element within the random hash is recorded.  Figure~\ref{fig:minhash} shows the example of a min-Hash computation, for three image signatures: $\{A,B,F\}$, $\{A,D,F\}$ and $\{B,C,E\}$, this results in an overall element vocabulary of $\boldsymbol{\nu} = [A,B,C,D,E,F]$, and  $\left\vert{\mathbf{N}}\right\vert=4$ random hash permutations, are formed. For the first random hash $\pi_1=\{F,B,A,E,D\}$ the index returned for SigA is 1 as the first index $F$ is present in the signature. For SigC, the index returned is 2 as the first index element, $F$, is not present in SigC but the second element $B$ is.

The similarity, $sim(\textbf{S}_{1},\textbf{S}_{2})$ is estimated as,
\begin{equation}
    sim(\textbf{S}_{1},\textbf{S}_{2}) = \frac{1}{\left\vert{N}\right\vert}\sum_{\forall \pi \in \textbf{N}} (min\ \pi (\textbf{S}_{1}) = min\ \pi(\textbf{S}_{2}))
\end{equation}
This means, the first matched index for each random hash are compared between signatures and the average number of identical pairs calculated. For the example in Figure~\ref{fig:minhash}, the estimated similarity between image signatures $A$ and $B$ will be 0.75 as they share 3 min-Hashes ($\pi_1$, $\pi_2$ and $\pi_4$), which is a close approximation to the exhaustive (naive) similarity. In contrast, the image signatures $A$ and $C$ share a single hash ($\pi_3$), giving an estimated similarity of 0.25.

By grouping the min-Hash results into ``sketches'', the false positive rate of the min-Hash is further reduced. A sketch is a grouping of min hash, where $K(S_{1})$ is the sketch $\{min\ \pi_{1}(S_{1}), ... , min\ \pi_{n} (S_{1})\} $ consisting of $n$ Hashes. A successful match between sketches is found if all the hash values are identical. By grouping the hashes, the false positive rate can be further reduced as similar image signatures will have many values of the min-Hash function in common and hence have a high probability of having the same sketches. On the other hand, unique image signatures will have a small chance of forming an identical sketch. 

In Figure~\ref{fig:minhash}, with a sketch size of $n=2$, the image signature sets $\{A,B,F\}$, and $\{A,D,F\}$  would be represented by the three sketches $\{(1,1)\}$, $\{(1,1)\}$ and $\{(1,2)\}$. Two out of the three sketches would match and therefore return a similarity of 2/3.


\subsection{Histogram weighting approximation}
\label{sec:Weighted}

\begin{figure}[htbp]
\centering
\includegraphics[width = 1 \columnwidth]{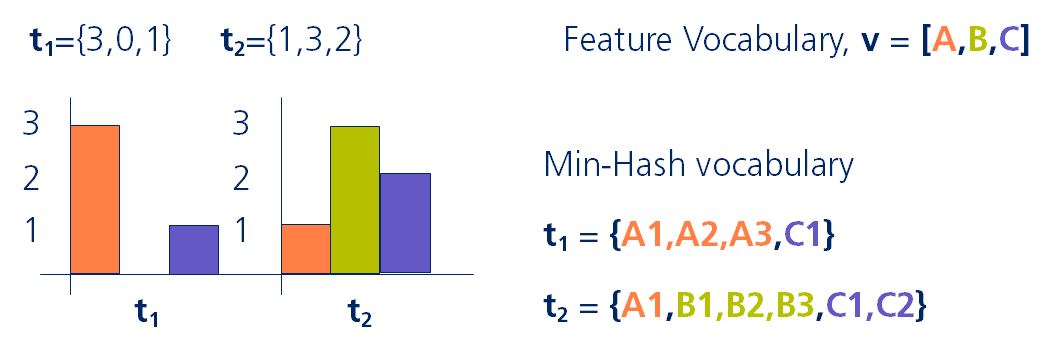}
\caption{The symbolisation of the image signature} \label{fig:WeightedminHash}
\end{figure}

The min-Hash algorithm assumes each element within a set is a unique "symbol" or element. However, an image signature is frequency based, so a new vocabulary has to be formed that accounts for the incidence of each item. Figure~\ref{fig:WeightedminHash} shows conversion of the frequency based histogram into related unique elements or symbolisation of the image signature.

For a visual vocabulary containing $|\boldsymbol{\nu}|$ visual words or features, for example $\boldsymbol{\nu} = \{A,B,C\}$, $\textbf{t}_i$ is a vector of the frequency response of the features. For example, with two input signatures, $\textbf{t}_{1}=\{3,0,1\}$ $\textbf{t}_{2}=\{1,3,2\}$, in order to convert the frequency based image signatures into a min-Hash based set of uniform symbols, the frequency of each feature in $\textbf{t}_{i}$ is used to duplicate symbols in $\textbf{t}_i'$. Therefore, in the example above, the min-Hash vocabulary for the two input signatures becomes, $\textbf{t}_{1}' = \{A1,A2,A3, C1\}$ $\textbf{t}_{2}' = \{A1,B1,B2,B3,C1,C2\}$. From this representation, the min-Hash method can be applied directly, where $sim(\textbf{t}_{1}',\textbf{t}_{1}')$ gives the pair wise similarity between the image signatures $\textbf{t}_{1}$ and $\textbf{t}_{2}$.

\section{Visualisation}
\label{sec:Greedy}

Min-Hash will return pairwise similarities between image signatures to present to the user via visualisation. We perform Agglomerative clustering between the resulting min hash values to emphasise distinct groupings in the data. For each signature, the closest signatures from the dataset are identified. They are said to be grouped if their similarity is greater than 66\%, and we repeat this process until no further grouping is possible.


Automatically grouping the data is effective in identifying similar content. However, as dataset size increases, it becomes increasingly difficult for the user to visualise effectively the groupings that emerge from clustering through textual methods alone. To overcome the visualisation challenge, multidimensional scaling (MDS) is used to visualise similarity regarding proximity in Euclidean space. MDS is a data analysis technique that displays the structure of distance-like data as a geometric picture. It was originally developed by Torgerson~\cite{TorgersonMDS}, in psychometrics to help understand people's judgements of the similarity of members of a set of objects.


MDS begins by constructing an initial configuration of the samples in the desired number of dimensions (generally 2 for this work).  This configuration is initially random and then iterates to convergence. Distances in the visualisation space are calculated with a Euclidean metric.   These distances are regressed against the original distance matrix. The predicted distances for each pair of samples are calculated, and the regression is by least-squares.  In a perfect visualisation, all visualised distances would fall exactly on the regression, that is, they would match the rank-order of distances in the original pairwise distance matrix from the min-Hash.  The \emph{goodness of fit} of the regression is measured based on the sum of squared differences between the visualisation-based distances and the distances predicted by the regression.  This \emph{goodness of fit} is called stress and is shown in equation~\ref{equ:Stress}. 
\begin{equation}
\label{equ:Stress}
    stress = \sum_{i=0}^{n}\sum_{j=i}^{n} (\left\| x_{i}-x_{j} \right\| - sim(\textbf{t}_{i}',\textbf{t}_{j}'))^2
\end{equation}
$i$ and $j$ are the possible samples, $x_{i} \in \textbf{x}$ is the sample in the Euclidean visualisation space, and $\textbf{t}_{i}'$ its signature. The objective of MDS is to optimise $\textbf{x}$ to minimise the deviation in this stress function. At each iteration, the positions of samples in visualisation space are moved by a small amount in the direction of steepest descent, i.e. the direction in which stress changes most rapidly. It is possible that local minima could occur, however given the MDS is only to help inform the user about the possible relationships between the media and is reformed at each iteration it is sufficient. The visualisation distance matrix is recalculated, the regression performed again, and stress recalculated. This entire procedure of nudging samples and recalculating stress is repeated until the procedure converges by failing to achieve any lower values of stress, which indicates that a minimum has been found. Effectively the visualisation means that two similar objects are represented by two points that are close together, and two dissimilar objects are represented by two points that are far apart. This allows the pairwise similarity and grouping of the data to be presented effectively to the user; the user is then able to select a single grouped subset and label only the small data subset, to identify co-occurring discriminative features through mining.
\section{Expanding signatures through co-occurring discriminatory features}
\label{ExpandSig}

Without learning, the MDS visualisation and groupings are purely based on the similarity of the initial image signatures which come from quantization of the feature space (e.g. a BoW). It is, therefore, unlikely that clustering and MDS will form meaningful groups. This is expected as there is often minimal inter (between) class variation, while lacking intra (within) class similarity. Therefore, we propose to ``push'' incorrectly labelled examples apart and to ``pull'' correct examples closer together. A variant of association rule data mining called APriori~\cite{Agrawal94}  is used to identify the compound elements from the signatures that are distinctive and descriptive within a subset of the correctly labelled examples when compared to the incorrectly labelled examples.  The new compound elements are then added to all the image signatures and this, in turn, will provide an increase in intra-class similarity. 


As we saw in Section~\ref{sec:Weighted}, given a feature vocabulary $\boldsymbol{\nu}$, any signature $\textbf{t}_i$ can be converted into a set of discrete symbols $\textbf{t}_i'$. In the language of association rule mining, the symbols are referred to as itemsets or transactions\footnote{The word \textbf{transaction} comes from the development of association rule mining in shopping basket analysis.} and the list of observed Transactions form a Transaction database, $\textbf{D} = \{\textbf{t}_{1}',...,\textbf{t}_{|\textbf{D}|}'\}$. The purpose of the APriori algorithm is to search this database and determine the most frequently occurring itemsets.

To achieve this efficiently, the APriori algorithm uses a bottom-up strategy to explore itemsets of increasing size. Initially single item itemsets are checked, and the itemset size is increased by one and this repeated. Only itemsets with a support and confidence greater than the threshold are retained. This allows the overall tree to be pruned to reduce the search space and makes the algorithm efficient when dealing with large itemsets.

An association rule of the form $\textbf{I}\Rightarrow \textbf{J}$ is evaluated by looking at the relative frequency of its antecedent and consequent parts i.e. the itemsets $\textbf{I}$ and $\textbf{J}$. The support of the itemset $\textbf{I}$ is the number of transactions in the overall database $\textbf{D}$ that contain $\textbf{I}$. The support of the rule $\textbf{I} \Rightarrow \textbf{J}$ is therefore,
\begin{equation}\label{SuportRuleAB}
   sup(\textbf{I} \Rightarrow \textbf{J}) = \frac{\left| \{ \textbf{t} \mid \textbf{t} \in \textbf{D}, (\textbf{I} \cup \textbf{J})\subseteq \textbf{t} \} \right|}{\left| \textbf{D} \right|}
\end{equation}
The support measures the statistical significance or importance of the rule, based on how often the rule occurs within $\textbf{D}$. However, the frequency of a rule across the dataset does not provide discriminative information. For multiple classes, discriminative rules are required. These are rules that occur within one class but not the others. To achieve this, the confidence of a rule is calculated as
\begin{eqnarray}\label{ConfRuleAB}
  conf(\textbf{I} \Rightarrow \textbf{J}) & = & \frac{sup(\textbf{I} \cup \textbf{J})}{sup(\textbf{I})}\nonumber \\
   & = & \frac{\left| \{ \textbf{t} \mid \textbf{t} \in \textbf{D}, (\textbf{I} \cup \textbf{J})\subseteq \textbf{t} \} \right|}{\left| \{ \textbf{t} \mid \textbf{t} \in \textbf{D}, \textbf{I}\subseteq \textbf{t} \} \right|} \nonumber \\
\end{eqnarray}

This means that the confidence is the ratio of the number of occasions when all the itemsets occur, relative to the number of cases in which the antecedent is present in the database. 

As an example, considering the vocabulary set of items $\boldsymbol{\nu}=\{A,B,C,D,E\}$, this might result in the following Transaction database, $\textbf{D}=\{\{A, B, C\},$ $\{A, B, C, E\},$ $\{A, B, E\},$ $\{A, C\},$ $\{A, B, C, D, E\} \}$ where $\left|\textbf{D}\right|=5$. The support of $(\{A,B\} \Rightarrow C)$ is 0.6 i.e. three occurrences of $\{A,B\}$ in five Transactions, while the confidence value is 0.5 i.e. two occurrences of $\{A,B,C\}$ in the four Transactions that contain $\{ A,B\}$.

To label a transaction as either a positive or negative class, the image signature $\textbf{t}_i'$ is appended with a label $\eta_i$, to mark it as a positive or negative example. The results of data mining then include rules of the form $\{A,B\} \Rightarrow \eta$ to give an estimate of $P(\eta | A,B)$ or the confidence of the association rule. $P(\eta | A,B)$ is only large and therefore used if $\{A,B\}$ occurs frequently in the positive examples but infrequently in the negative examples.  If $\{A,B\}$ occurs frequently in both positive and negative examples i.e. several classes, then $P(\eta | A,B)$ will remain small as the denominator in equation~\ref{ConfRuleAB} will be large.  The confidence threshold is set to 1, to ensure that association rules are only found if the elements are contained in the positive set and none of the negative sets.
\section{Iterative signature learning}
\label{IterSig}

Association rule mining is performed on a selected subset of positive and negative image signatures, but the resultant rules are applied to all the signatures in the dataset. For each rule returned from the mining, all signatures are searched for the occurrence of that rule. Depending on whether the operation seeks to increase similarity (pull together) or dissimilarity (push apart), an additional element is added or removed respectively. 

For example, if the rule returned a single element (A2), this relates to the feature $A$, and given the image signature $\textbf{t}_{1}' = \{A1,A2,A3, C1\}$, an additional element related to the $A$ feature, element $A4$ would be added. If the rule returned had multiple items for example (A2, B6), and joint feature $AB$ would be added to the image signature. However if the image signature doesn't contain any $A$ features, no additional elements would be added. This increased weighting on the feature $(A)$ would ``pull" together sets that contain $(A)$ features over time improving accuracy. In addition, the mining can return association rules that contain multiple subsets that together are descriptive and distinctive. Using the same example, if the mining returns the rule $(A1, B1)$, the compound feature $AB1$ would not be appended to $\textbf{t}_{1}'$ as the set does not contain any $(B)$ features. However, given the image signature $\textbf{t}_{2}' = \{A1,B1,B2,B3,C1,C2\}$, the compound feature $AB1$ would be appended making $\textbf{t}_{2}'= \{A1,B1,B2,B3,C1,C2,AB1\}$. Thereby increasing the importance of the co-occurrence of the $A$ and $B$ features together.

In contrast, if a \emph{push apart} operation is performed, for each rule returned from the mining, if the image signature contains the elements of the rule, the min-Hash element would be \emph{removed}. This would reduce similarity between the correct and incorrect image signatures resulting in items being pushed apart in the MDS space. Using the sample example above, if the association rule returned from the mining highlighted feature $(A)$, the element $(A3)$ would be \emph{removed} from set $\textbf{t}_{1}'$ and the element $(A1)$ would be removed from set $\textbf{t}_{2}'$. This would reduce the set overlap between the correct and incorrect image signatures to ungroup them in the MDS visualisation.

The min-Hash distance computations, clustering and grouping process can then be repeated and the MDS visualisation redrawn to illustrate the improved grouping of the media.
\section{Results}
\label{sec:Results}
To illustrate the approach and evaluate the quality of the clustering and categorization, testing is performed on a variety of datasets. We report results on image, text and video using both feature detectors and classifier outputs to show the generality of approach. Furthermore, all results reported use a nearest neighbour classifier to achieve state-of-the-art performance which demonstrates the power of the learning approach. Datasets include the 15 Scene dataset~\cite{LazebnikCVPR06}, the caltech101~\cite{FeiFeiCVPR04} image datasets, FG-NET~\cite{panis2015overviewFG-NET} a human age image dataset, a mixed media dataset ImageTag~\cite{GilbertACCV12}, the video datasets KTH~\cite{Schuldt04}, UCF11~\cite{LiuCVPR09} and Hollywood2~\cite{MarszalekCVPR09}, and the active learning dataset Iris~\cite{Lichman2013}. Table~\ref{tab:datasets} gives an overview of the properties of the datasets used, although it should be noted that due to the online/active learning not all the labelled training data is used.
 
\begin{table}[htbp]
\caption{Dataset properties}
  \begin{center}
\begin{tabular}{|c||c|c|c|c|}
  \hline
Dataset                                                        &    Mode            & Num                 &Num                     &Num            \\    
                                                                    &                        &    Class            &    Train        &    Test            \\ \hline
15Scene~\cite{LazebnikCVPR06}            &Image            & 15                    &    1500                & 2986            \\
Caltech101~\cite{FeiFeiCVPR04}        &Image            & 101                    &    1515                & 4040            \\
FG-NET~\cite{panis2015overviewFG-NET}    &Image            &    -                        & 300                    & 601                \\
ImageTag~\cite{GilbertACCV12}            &Img+Txt        & 14                    & 2334                &  466             \\
KTH~\cite{Schuldt04}                            &Video            &    6                        &    192                    & 192                \\
UCF11~\cite{LiuCVPR09}                    &Video            & 11                    &    974                    & 194             \\        
Hollywood2~\cite{MarszalekCVPR09}    &Video            & 12          &    810                    & 810                \\     
Iris~\cite{Lichman2013}                    &Num                & 3                        &    150                    & 150                \\  \hline                                              
\end{tabular}
\end{center}
\label{tab:datasets}
\end{table}

\subsection{Evaluation} 

To evaluate the success of our approach, we use two types of validation measures: Classification performance, taking individual group cluster means and comparing to a ground truth class label, this classification performance can be obtained using a nearest neighbour classifier for standard comparison to other approaches. Also, we examine cluster purity, the purity of a group is given by
\begin{eqnarray}\label{ClusterPurity}
  purity(\Omega,\textbf{C}) = \frac{1}{N}\sum_{k}\stackrel{max}{j}|\omega_{k}\cap c_{j}|
\end{eqnarray}
where $\Omega = [\omega_{1}, \omega_{2}, ..., \omega_{k}]$ is a set of the groups and $\textbf{C} = [c_{1}, c_{2}, ..., c_{j}]$ is the possible class labels. In general, the larger the value of purity, the better the solution. Purity is limited if the number of clusters is high, however, in our case the number of groups is low relative to the size of the data. All datasets have internal validation, and a more in-depth examination of cluster purity is carried on the video datasets UCF11 and Hollywood2.

\subsection{Image Datasets}

First, we evaluate on three pure image datasets.

\subsubsection{15 Scene Dataset}
The 15 Scene category dataset by Lazebnik~\cite{LazebnikCVPR06} consists of 4486 grey scale images across 15 classes such as \emph{kitchen, industrial, tall building and street}. Forming the original image signatures from a 512 dimensional GIST~\cite{OlivaIJCV01GIST} feature vector computed for each image. For this example, the GIST feature vector is normalised and then directly converted into a symbolised signature as described in Section~\ref{minHashAlgorithm} i.e. there is no BoW employed. To allow comparison with other approaches, the training/test partitioning proposed by Lazebnik~\cite{LazebnikCVPR06}, was used. From each class, 100 images are selected for training, providing a pool of up to 1500 training images, retaining the remaining images for testing only, image signatures were formed for all 4486 images. At each iteration of learning, ten images are selected from the training pool (9 correct examples from one of the classes and one incorrect from another class:$<$9T,1N$>$) from the MDS visualisation and clustering of the signatures. All signatures are then adapted according to the rules identified during mining. The objective of which is to try and pull the correct examples together. Experiments are repeated for ten different random training partitions, and the mean accuracy and standard deviation reported. It is important to note that while 1500 training images are available to the nearest neighbour classifier, the learning approach can achieve state-of-the-art performance using only 180 of these images. The results are shown below in Table~\ref{tab:15scene}.

\begin{table}[htbp]
\caption{Accuracy of 15 Scene dataset $<$9T,1N$>$}
  \begin{center}
\begin{tabular}{|c||c|c|c|}
  \hline
                Approach                                                                                                                                                            &  Accuracy       &Train GT &  $\sigma$ \\
                                                &                 & Imgs Used  &     -  \\ \hline   
                Lazebnik~\cite{LazebnikCVPR06}  &    81.4\%       & 1500       &     -  \\ 
                Nakayama~\cite{NakayamaCVPR10}  &    86.1\%       &  1500      &     -  \\ \hline \hline 
                Iter 0 Baseline                 &    10.47\%      &  0         &     3.7\%  \\
                Iter 1                          &    24.59\%      &  10        &     18.9\%  \\
                Iter 2                                                                                                     &    41.89\%                          &  20        &     21.8\%  \\
                Iter 4                                                                                                     &    60.50\%                          &  40        &     24.6\%  \\
                Iter 6                                                                                                     &    62.78\%                          &  60        &     25.0\%  \\
                Iter 8                                                                                                     &    65.12\%                          &  80        &     22.9\%  \\                                                
                Iter 10                                                                                                   &    70.16\%                          &  100       &     25.1\%  \\
                Iter 12                                                                                                   &    71.58\%                          &  120       &     14.1\%  \\
                Iter 14                                                                                                   &    78.59\%                          &  140       &     6.9\%  \\
                Iter 16                                                                                                   &    81.24\%                          &  160       &     4.7\%  \\                                 
                Iter 18                                                                                                   &    89.87\%                          &  180       &     2.6\%  \\
                Iter 20                                                                                                   &    \textbf{89.91}\%                          &  200       &     2.05\%  \\ \hline                                                 
                                
\end{tabular}
\end{center}
\label{tab:15scene}
\end{table}

The results show that initially (grouping on raw GIST features), the performance is low but increases dramatically with each iteration of the algorithm. The user is effectively training each class in turn by identifying and removing confusion within classes. Given the 15 classes it might be expected that 15 iterations would be required and table~\ref{tab:15scene} supports this with no further accuracy gains after the 18\textsuperscript{th} iteration (180 training images). At this point, accuracy levels surpass the state-of-the-art while using considerably less data. This is expected as the user is heavily involved in the training process, allowing common confusion areas to be identified and removed. 

\begin{figure}[htbp]
\centering
\includegraphics[width = 1 \columnwidth]{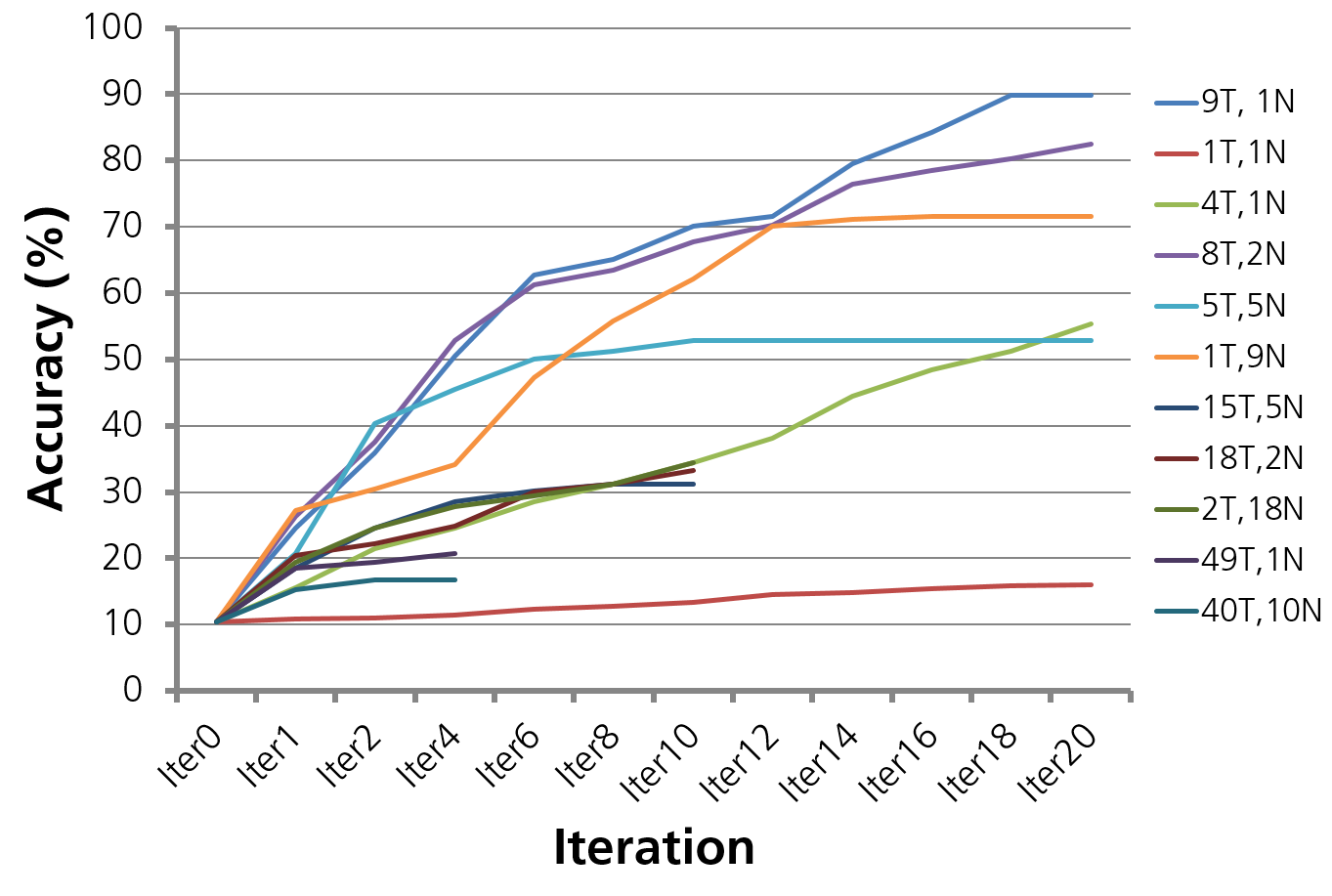}
\caption{15 Scene Accuracy for different ratios of correct and incorrect selected subsets}
\label{fig:15SceneAccGraph}
\end{figure} 
Figure~\ref{fig:15SceneAccGraph} shows the effect on the accuracy of choosing different ratios of correct vs. incorrect examples at each iteration. The ratio of correct vs. incorrect selected examples of $<$9T,1N$>$ and $<$8T,2N$>$ perform the best gaining state of the art performance after 20 iterations with an accuracy of 89.91\% and  82.5\% respectively (see table~\ref{tab:15scene}). The ratio $<$4T,1N$>$ performs well over the 20 iterations but at about half the performance of $<$9T,1N$>$, which is expected. Some combinations result in fewer rules extracted during mining, and this increases the number of iterations that are required. However, it is encouraging that all combinations increase accuracy with each iteration and the choice of subsets size only effects speed of convergence. When the performance no longer increases, this is due to no new co-occurring mined rules being identified. Therefore, no changes are made to the image signature, thus ensuring that the approach does not overfit.

\subsubsection{Caltech101}

  To provide a more challenging test, and demonstrate
flexibility to features, the commonly used benchmark
dataset, Caltech101~\cite{FeiFeiCVPR04} was also evaluated. The dataset
consists of 101 object categories with between 31 to
800 images per category, using 15 training examples
randomly selected from each class, forming a training pool of 1515 images that the user could select from.
Two descriptors were applied to the dataset, SIFT and
CNN features. The SIFT-based image signatures were
formed from a 512 element BoW histogram of standard
SIFT descriptors after they have been reduced
to 30 dimensions through PCA as employed in~\cite{CaiPAMI10}.
After symbolisation, the average image signature for
this dataset is of size 2150. The CNN features are
extracted from a deep CNN model pre-trained with
the ImageNet dataset. We extract features from the
sixth layer of the network which has the same architecture
as that proposed by~\cite{krizhevsky2012imagenet} and won ILSVRC2012.
Because deep CNN-based features are extracted from
the network, which is trained for recognition tasks, we
can regard it as a feature that expresses discriminative
information of an image, in our tests, we use the Caffe
implementation~\cite{jia2014_caffe}. We convert this 4096 feature
response directly to an image signature, rather than
using a codebook.
The image signature for each feature descriptor is
iteratively adapted with the ratio $<$4T,1N$>$ at each iteration. Classification performance is evaluated on 40
unseen ”test” images from each class by performing
a nearest neighbour assignment to the closest class.
The experiment was repeated for ten user runs using
different training/test partitions with each user performing
15 iterations during learning. Average results
and the standard deviation $\sigma$ are shown in table~\ref{tab:caltechResults}.

\begin{table}[htbp]
\caption{Accuracy of Caltech101 dataset}
  \begin{center}
\begin{tabular}{|c||c|c|c|}
  \hline
                Approach                                                                                                                                                           &  Accuracy       &Train GT &  $\sigma$  \\
                                &                 &  Imgs Used &     -          \\ \hline   
Cai~\cite{CaiPAMI10}            &    64.9\%       & 3030       &     -          \\ 
Wang~\cite{WangCVPR10_Caltech}    &         65.4\%                &    3030             &         -            \\
Sohn~\cite{SohnICCV11}                        &         71.3\%                &    3030             &          -           \\  
Chatfield~\cite{Chatfield14_CNN}&         88.5\%                &    3030             &         0.33\%    \\ \hline \hline        
SIFT Iter 0                     &    21.54\%      &  0         &     2.1\%  \\
SIFT Iter 1                     &    32.78\%      &  10        &     10.8\% \\                  
SIFT Iter 5                     &    41.62\%      &  50        &     18.8\% \\                     
SIFT Iter 10                    &    51.2\%       &  100       &     7.8\%  \\              
SIFT Iter 15                    &    59.7\%       &  150       &     3.8\%  \\ \hline \hline
CNN  Iter 0                     &    45.8\%      &  0         &     1.9\%  \\
CNN  Iter 1                     &    52.1\%      &  10        &     10.4\% \\                  
CNN  Iter 5                     &    75.8\%      &  50        &     18.7\% \\                     
CNN  Iter 10                    &    84.1\%       &  100       &     2.9\%  \\              
CNN  Iter 15                    &    \textbf{89.7\%}&  150       &     0.7\%  \\ \hline                                                                  
\end{tabular}
\end{center}
\label{tab:caltechResults}
\end{table}
Iterative learning demonstrates large performance
increases over the baseline signature for both feature
types. The approach by Cai~\cite{CaiPAMI10} (64.9\%), compares
well with iterative learning with SIFT features at
59.7\% but our approach also demonstrates similar
performance (89.7\%) to that of Chatfield~\cite{Chatfield14_CNN} at 88.5\%
using the CNN features. In both cases considerably
fewer examples are used in training. While the CNN
features themselves were trained on a much larger
dataset, the outcome substantiates our claims to the
flexibility of the learning framework and demonstrates
that the limiting factor in learning is actually
the feature representation, not the approach.

\subsubsection{Relative Human age prediction}
We use the FG-NET image age dataset~\cite{panis2015overviewFG-NET} to predict the relative attribute of comparative human age between pairs of images. Given the challenge of the subjective nature of age prediction even for humans~\cite{fu2010}, this is an interesting avenue for our approach given the integral nature of humans in our learning process. The FG-NET dataset consists of 1002 images of 82 individuals labelled with ground-truth ages ranging from 0 to 69. To follow and compare to work of~\cite{Fu2015Fgnet} we used up to 300 images for training and the remainder for the test. The experiments were repeated ten times, and each image is represented by a 200 dimension AAM vector. Pairwise comparisons were formed using the data collected from an online study~\cite{Fu2015Fgnet}. A total of 4000 pairwise image comparisons were collected from 20 participants and given that human can be error prone for small ages differences these comparisons contained \emph{unintentional errors}. To demonstrate the strength of the approach to mislabelled data, we generate additional \emph{intentional errors}. These are introduced by using additional random image comparisons. Note that for this dataset the MDS visualisation was adapted as the human image comparisons were already collected. To rank the results of the approach, MDS was applied to the pairwise similarity matrix of the test dataset, with a dimension of 1, to create a ranked list of images of increasing human age. Random pairwise labelled image comparisons from the training set were iteratively compared to the MDS resultant ranked list of images to identify incorrectly ranked image pairs. The incorrectly ranked image pairs were  used as the true and false selections to adapt the image signatures. To compare against other approaches the Kendall tau rank correlation was used. We quantitatively compared our approach against three methods; \emph{URLR}~\cite{Fu2015Fgnet}, a joint ranking and outlier learning method; \emph{Huber-LASSO}~\cite{xu2013robust}, a statistical ranking method that performs outlier detection and \emph{GT}, the upper bound of the training data. 
\begin{table}[htbp]
\caption{FG-NET  dataset with Unintentional errors}
  \begin{center}
\begin{tabular}{|c||c|c|c|c|c|}
  \hline
                                                                                                                                                                          
   \%    Train                               &   GT           &  URLR       & Huber             & Sig            &$\sigma$ Sig\\   
    images                                                 &                                &                            & LASSO             &    min-H            &min-H \\ \hline
                            0                                    &        0.686                    &        0.651         &  0.651             &    0.552                    & 0.01\\
                            10                                 &  0.686                     &    0.686            & 0.675                & 0.680                    & 0.11\\
                            20                                &        0.686                    & 0.680                & \textbf{0.678}&    0.681                    & 0.06\\
                            30                                 &    0.686                    &    \textbf{0.682} &    0.670                & \textbf{0.685}& 0.02\\
                            40                                 &    0.686                    &    0.680                &    0.671          & \textbf{0.685}&0.00\\
                            50                                    & 0.686                    &    0.681                & 0.668              & \textbf{0.685}&0.00\\ \hline        
\end{tabular}
\end{center}
\label{tab:fg-net}
\end{table}

\begin{figure}[htp]
\centering
\includegraphics[width = 1 \columnwidth]{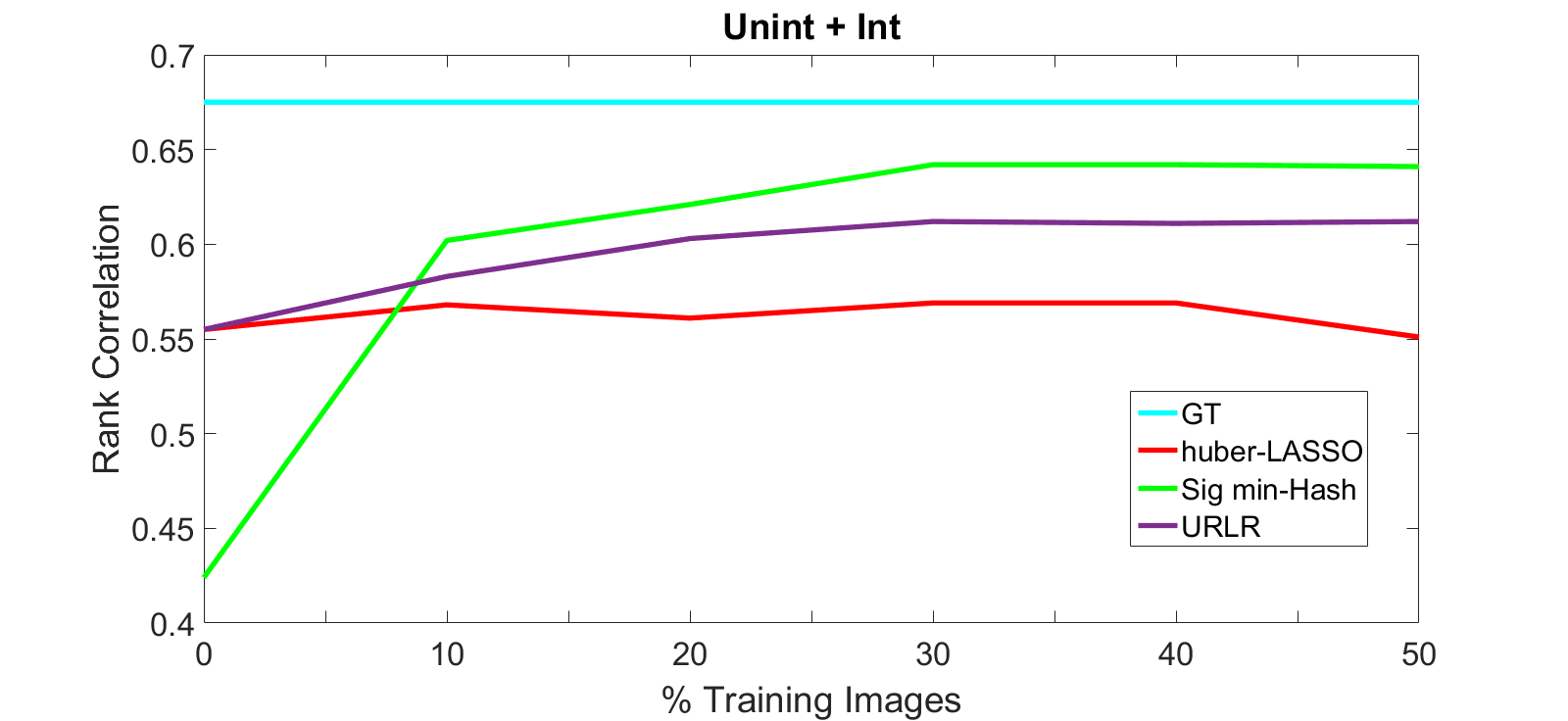}
\caption{Comparing our Sig min-Hash with URLR and Huber-LASSO with Unintentional+Intentional errors}
\label{fig:FGNetIntentGraph}
\end{figure}

\begin{table}[htbp]
\caption{FG-NET  dataset with Unintentional+Intentional errors}
  \begin{center}
\begin{tabular}{|c||c|c|c|c|c|}
  \hline
                                                                                                                                                                          
   \%    Train                               &   GT           &  URLR       & Huber             & Sig            &$\sigma$ Sig\\   
    images                                                 &                                &                            & LASSO             &    min-H            &min-H \\ \hline
                            0                                     &        0.675                 &        0.555         &  0.555              &    0.424                & 0.07    \\
                            10                                 &  0.675                     &    0.583            & 0.568                & 0.602                  & 0.13\\
                            20                                 &        0.675                 & 0.603                & 0.561                &    0.621                & 0.09    \\
                            30                                 &    0.675                     &    0.612                &\textbf{0.569}                & \textbf{0.642}& 0.04    \\
                            40                                 &    0.675                     &    0.611                &\textbf{0.569}                & \textbf{0.642}& 0.01    \\
                            50                                 & 0.675                     &\textbf{0.612}& 0.551                &    0.641                & 0.00    \\ \hline        
\end{tabular}
\end{center}
\label{tab:fg-net+Intent}
\end{table}
All methods are robust to the low unintentional error in table~\ref{tab:fg-net} with a performance close to the ground truth after minimal training data is used. However when the training data is corrupted with intentional errors, table~\ref{tab:fg-net+Intent} and figure~\ref{fig:FGNetIntentGraph} demonstrate the effective performance of our proposed method \emph{Sig min-Hash} at being significantly better than URLR and Huber-LASSO. This is due to the online learning adapting the image signatures of the data in an iterative process, but using the mining only to identify common descriptive rules between image signatures, and therefore not corrupting the image signature with intentional errors. 

\subsection{Mixed Media ImageTag Dataset}
\begin{figure*}[htp]
\centering
\includegraphics[width = 1.8 \columnwidth]{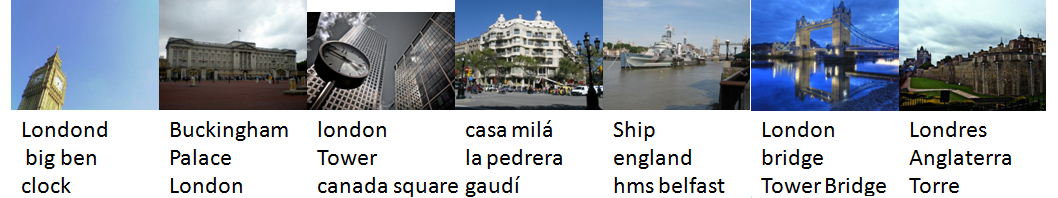}
\caption{Example image and tags from the \emph{ImageTag} dataset}
\label{fig:ImageTagSamples}
\end{figure*}

As an extension to the image datasets, the approach is also tested on the ImageTag dataset~\cite{GilbertACCV12}. ImageTag contains 2800 images and associated meta-data (tags) from the internet image site \emph{Flickr}. 
It consists of 14 classes of tourist sites in both London and Barcelona with 200 images per class. The sites are: \emph{Big Ben, Buckingham Palace, Canada Square, Casa Mila, HMS Belfast, London Bus, Sagrada Familia, St Pancras, St Pauls, Torre Agbar, Tower Bridge, Tower of London, Wembley, Westminster Abbey}. 
Figure~\ref{fig:ImageTagSamples} gives examples of the images and some of the tags. The tags are missing from around 50\% of the images, and can contain foreign languages, and spelling mistakes. 
Due to the use of the image signature container any tags from the metadata can be concatenated to the image features for each piece of the media, boosting the performance by combining both the text and image features.
Each image is described by a visual Bag of Words
(BoW) histograms of standard SIFT descriptors with
the dimension reduced to 30 as in the previous section.
A BoW histogram is also built for the textual tags and concatenated to the visual BoW to form the initial image signature. There are 197 textual
labels and initially 9053 unique symbols from the SIFT
descriptors. Repeating the experiments for 20 user runs with 20 iterations of learning per run. Table~\ref{tab:ImageTagPerf} shows the performance of the image signature formed
of only the image SIFT descriptors, the text tags
(only images with tags are included in the test) and
combined image and textual descriptors.

\begin{table}[htbp]
\caption{Accuracy of ImageTag dataset}
  \begin{center}
\begin{tabular}{|c||c|c|c|c|} \hline
\multirow{2}{*}{}    & SIFT    &\multirow{2}{*}{Text tag}     & SIFT    & $\sigma$ SIFT+   \\ 
                     & Desc    &                              &  + TextTag   &    TextTag           \\ \hline                    
Iter 0                                     &   25\%       &   45.9\%                   & 28.75\%      &     2.3          \\ 
Iter 1                    &    27.5\%    &  60.2\%                    & 32.2\%       &      13.8        \\                                      
Iter 5                    &    43.7\%    &\textbf{61.4\%}& 54.75\%      &      9.4         \\ 
Iter 10                   &   65.4\%     & 60.9\%                     & 72.4\%       &      4.2         \\       
Iter 20                   &\textbf{69.4\%}& 61.2\%                     &\textbf{73.4\%}&      1.8         \\ \hline       
\end{tabular}
\end{center}
\label{tab:ImageTagPerf}
\end{table}

The initial baseline performance is shown as Iter 0, 
the accuracy increases sharply over the 20 iterations. It can be seen that the combination of the text tags and SIFT image descriptors increases the accuracy. This is expected due to the quality of the tags, but considering only 50\% of the images are tagged, these results are encouraging. 

\begin{figure}[htp]
\centering
\includegraphics[width = 1 \columnwidth]{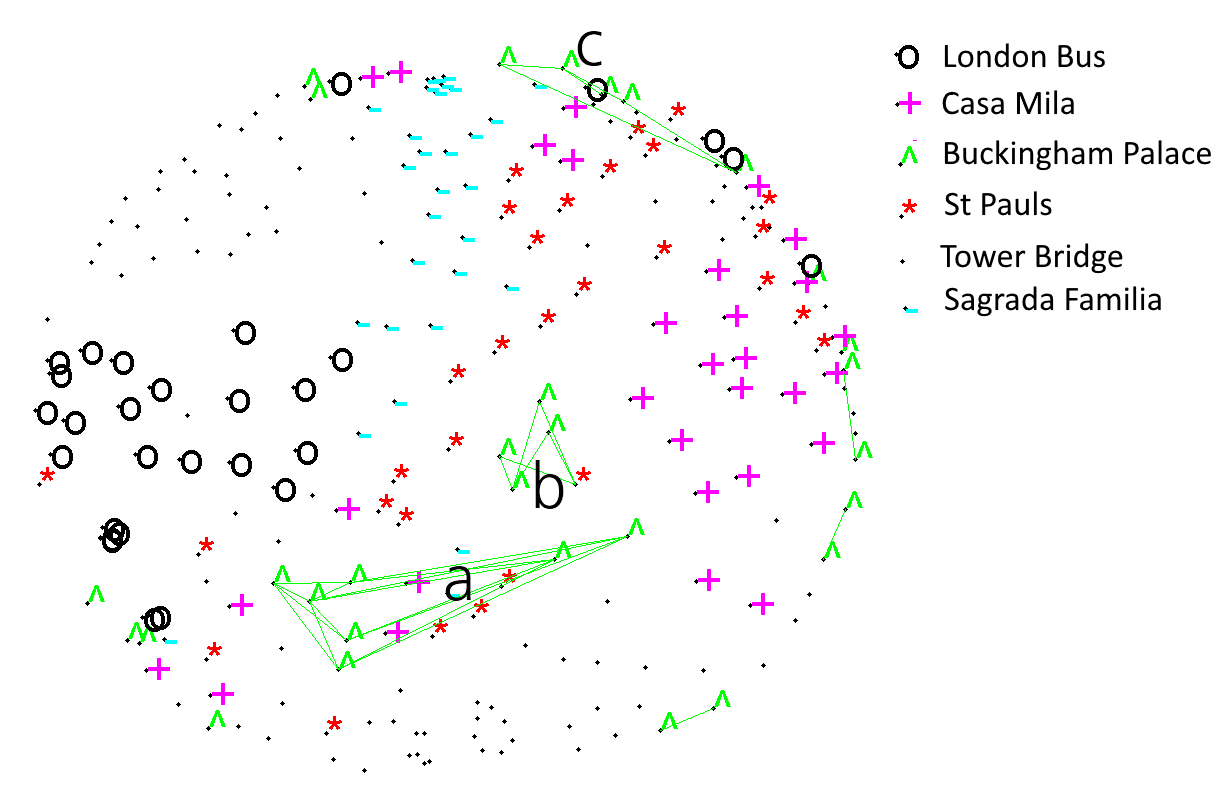}
\caption{Figure~\ref{fig:GroupBuckingham} shows the MDS visualisation of the ImageTag dataset after 10 iterations. Each symbol represents a different class, and Euclidean distance indicates similarity. Also, the grouping of the Buckingham Palace class (green hat symbols) can be seen, by the lines. There are some incorrectly grouped images within area a) ( the plus and stars), however, most of the groups are correct.}
\label{fig:GroupBuckingham}
 \end{figure}

Figure~\ref{fig:GroupBuckingham} shows the grouping after ten iterations. The groupings are formed from the agglomerative clustering, as can be seen, the groups are relatively distinct. 

\begin{figure}[htp]
    \centering
    \subfigure[]{
    \includegraphics[width = 0.29 \columnwidth]{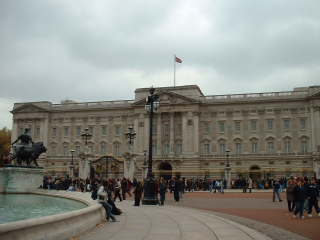}
    \label{fig:BuckinghamPalceSubGroups_a}}
    \subfigure[]{
    \includegraphics[width = 0.29 \columnwidth]{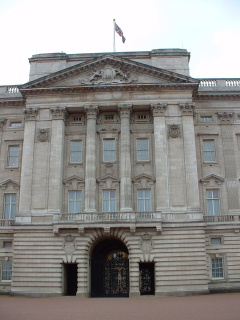}
    \label{fig:BuckinghamPalceSubGroups_b}}
        \subfigure[]{
    \includegraphics[width = 0.29 \columnwidth]{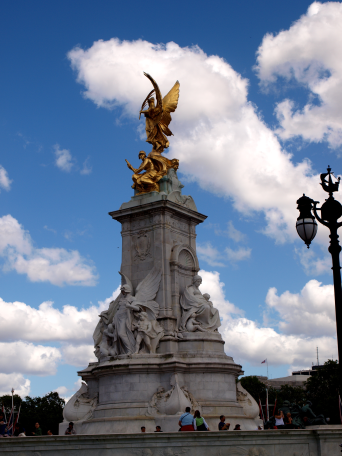}
    \label{fig:BuckinghamPalceSubGroups_c}}
    \caption{Representative images from the three lettered sub groups in Figure~\ref{fig:GroupBuckingham}, a, b and c}
    \label{fig:BuckinghamPalceSubGroups}
\end{figure}
For the class Buckingham Palace (marked by the green "hat" symbols), region \emph{c)} is quite distant to regions \emph{a)} and \emph{b)}. However, as can be seen in Figure~\ref{fig:BuckinghamPalceSubGroups_a} and~\ref{fig:BuckinghamPalceSubGroups_c} there is a large visual difference between these groups and would therefore be expected.

\subsection{Video datasets}
Due to the generic and efficient design of our online learning approach, it is well suited to large video media datasets such as the UCF11 dataset~\cite{LiuCVPR09} or Hollywood2~\cite{MarszalekCVPR09}. We demonstrate applying online learning using two different feature approaches; applying the 2D compound corners of Gilbert \emph{et al}~\cite{GilbertICCV09} to the KTH~\cite{Schuldt04} and UCF11~\cite{LiuCVPR09} datasets and the dense trajectories of Wang \emph{et al}~\cite{wang2013denseTraj} to the  Hollywood2~\cite{MarszalekCVPR09} dataset.

\subsubsection{KTH Dataset}
\label{sec:KTH}
 The KTH dataset~\cite{Schuldt04} contains 6 different actions; \emph{boxing, hand-waving, hand-clapping, jogging, running} and \emph{walking}.
There is a total of 25 people performing each of the 6 actions, four times; giving 599 video sequences (1 sequence is corrupt). Each video contains four instances of the action totalling 2396 individual actions. We present results using training and test partitions as suggested
by Sch\"{u}ldt~\cite{Schuldt04}, with eight people for training, and eight people for testing. The features are formed on the training subset
using the approach by Gilbert~\cite{GilbertICCV09} where the features
consist of compound corner classifiers. The compound corners are the result of learnt hierarchically
grouped 2D Harris corners in space and time that represent a spatiotemporal structure that is indicative of specified actions, with a separate classifier learnt for each action class. The image signature consists of the six classifiers concatenated, with the original image signature containing 1204 unique symbols, formed from the
frequency count of each compound corner symbolised to provide the original signature. The experiment was repeated for 20 user runs with 10 iterations of learning per run. At each iteration, the
true and negative selection of the videos is ($<$5T,1N$>$).
Figure 7 shows the MDS visualisation after only 60 labelled videos for the class, handclapping. The videos
are well grouped despite being spatially close to other
classes. Also, it is interesting to see the separation
between the first three static classes, boxing,
handclapping and handwaving (the pink cross, red
star and green hat) at the top of the image, and the
dynamic classes, jogging, running and walking in the
lower part of the picture.
\begin{figure}[htp]
\centering
\includegraphics[width = 1 \columnwidth]{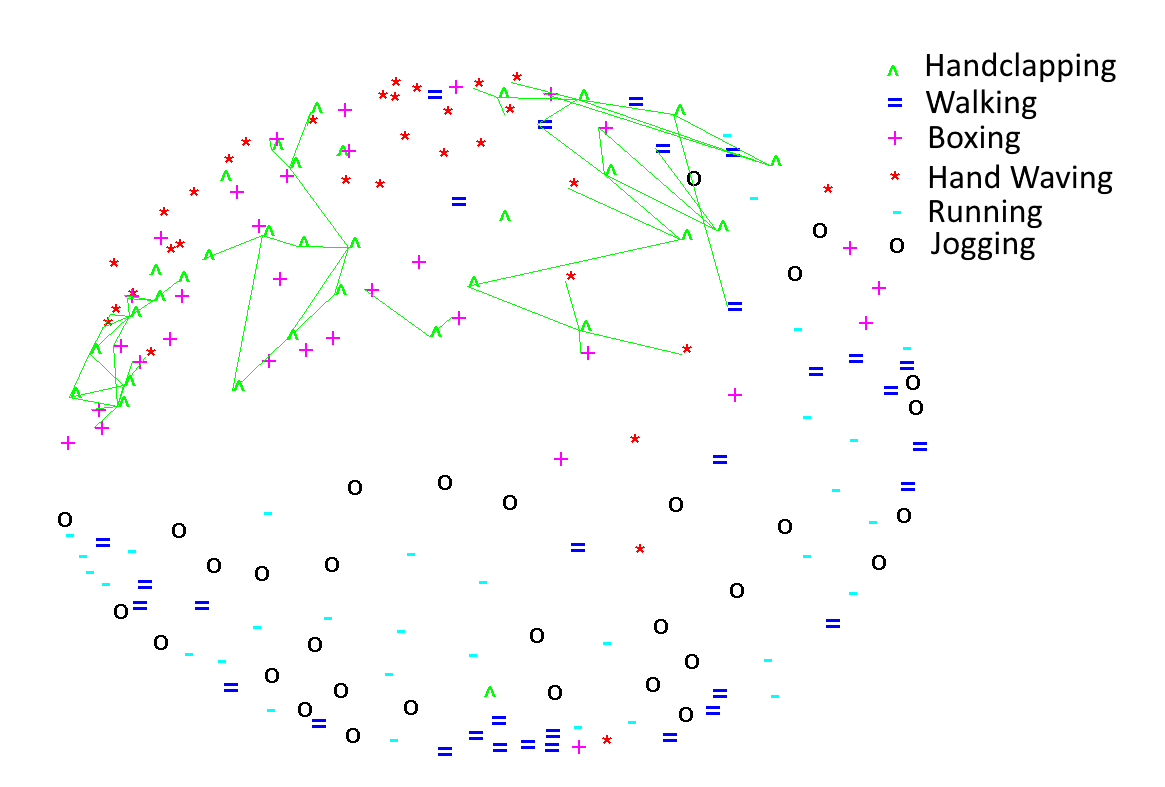}
\caption{MDS Visualisation of the grouping of the Handclapping class from the KTH Dataset after ten iterations, each class is indicated by a different symbol}
\label{fig:GroupKTH}
\end{figure}

The accuracy for up to 60 labelled videos with ten iterations is 91.2\% this compares well with the baseline min-Hash of 44.3\%. Furthermore, Table~\ref{tab:KTHAcc} shows the results compared to other approaches and despite the mined min-Hash approach only needing 42 labelled videos, the accuracy is comparable to the state of the art approaches using the traditional train/test method, but with $\frac{1}{5}$ of the labelled training data.
\begin{table}[htbp]
\caption{Accuracy on the KTH dataset}
  \begin{center}
\begin{tabular}{|c||c|c|c|c|}        
  \hline
   Approach                              &  Iter          &Acc       & Train GT         &$\sigma$    \\ 
                                         &         -  &              & Vids Used    &    -                \\ \hline
   Sch\"{u}ldt~\cite{Schuldt04}          &         -  &  71.71\%     &  192                  &    -                \\
   Klaser~\cite{KlaserBMVC08}            &           -  &  91.5\%      &  192                  &    -                \\
   Laptev~\cite{LaptevCVPR08}               &         -  & 91.8\%       &  192                  &    -                \\                
   Wang~\cite{WangBMVC09}                &         -  &  92.1\%      &  192                  &    -                \\ 
   Kovashka~\cite{KovashkaCVPR10}        &         -  &  94.5\%      &  192                  &    -                \\
   Gilbert~\cite{GilbertICCV09}          &         -  &  \textbf{95.7\%}&  192                  &    -                \\\hline \hline
   Baseline                              &         0  &   44.3\%     & 0                     &    1.3\%        \\ 
   Sig min-Hash                                &        2   &   61.4\%     & 12           &    20.5\%  \\ 
   Sig min-Hash                                &        5   &   80.7\%     & 30           &    11.9\%  \\ 
   Sig min-Hash                                &        7   &   91.2\%     & 42           &    3.2\%   \\                            
   Sig min-Hash                                &        10  &   91.2\%     & 60           &    0.6\%      \\ \hline 
\end{tabular}
\end{center}
\label{tab:KTHAcc}
\end{table}

\subsubsection{UCF11 Dataset}
The YouTube based dataset, UCF11~\cite{LiuCVPR09} consists of eleven categories: \emph{basketball shooting, cycling, diving, golf, horse riding, juggling, play swings, tennis swinging, trampolining, volleyball, and dog walking}.  The videos are all captured from videos uploaded onto the YouTube website, consisting of 1168 videos that exhibit large variations in camera motion, object appearance and pose, object scale, viewpoint, cluttered background and illumination conditions. 
%
%
%
The feature descriptor for the UCF11 dataset is the compound corner features classifiers trained on the KTH dataset (see section~\ref{sec:KTH})\footnote{To allow other to make comparison the feature responses for the UCF11 dataset are made available here www.andrewjohngilbert.co.uk/features.html}.

The 6 KTH action classifiers are concatenated into a single vector, the frequency count of each compound corner on the UCF11 video recorded and symbolised to provide an initial signature for the UCF11 dataset. 

\begin{figure}[htp]
\centering
\includegraphics[width = 1 \columnwidth]{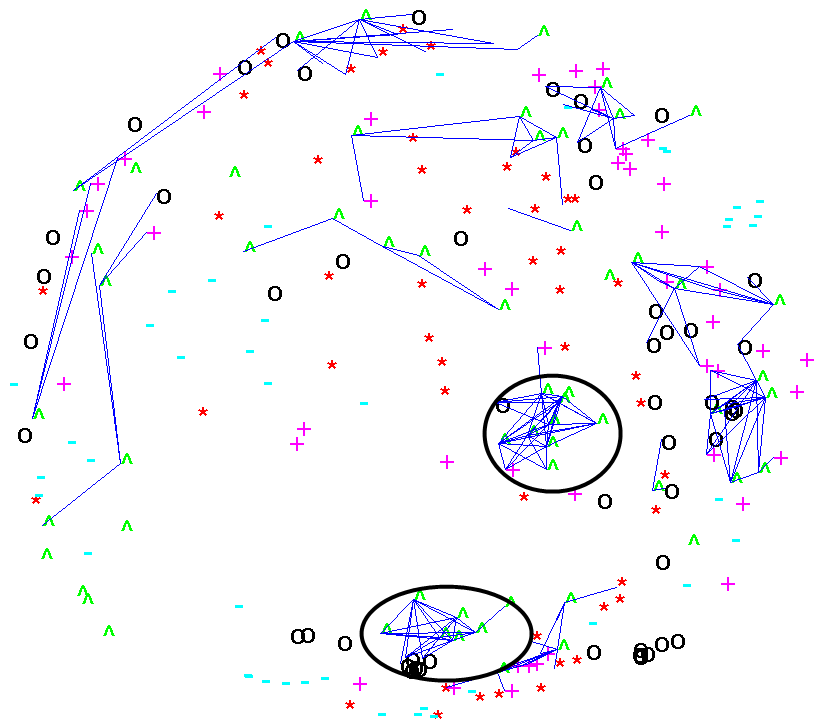}
\caption{Initial greedy clustering result of the class Diving, indicated by the green hat symbol, 2 groups of the class diving are indicated by the black circles}
\label{fig:InitalGroup_diving_Iter1}
\end{figure}

The image signature for each video contains around 2000 elements and the total number of unique elements, or the initial vocabulary is 3108 elements. Figure~\ref{fig:InitalGroup_diving_Iter1} shows the initial groupings for the class \emph{Diving} from the UCF11 dataset, where each symbol represents a different class.

It can be seen that there are a number of groups of correct examples but also many incorrect examples. Overall for the UCF11 dataset, there are initially 60.4\% correct groupings and 21.4\% incorrect groupings.
\begin{figure}[htp]
    \centering
    \subfigure[Initial greedy clustering result of the class Diving]{
    \includegraphics[width = 0.45 \columnwidth]{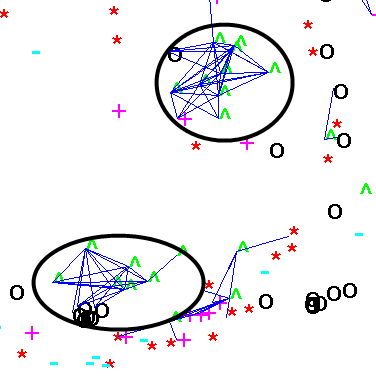}
    \label{fig:InitalGroup_divingcrop}}
    \hfil
    \subfigure[Grouping of diving class after pulling together two separate groups]{
    \includegraphics[width = 0.45 \columnwidth]{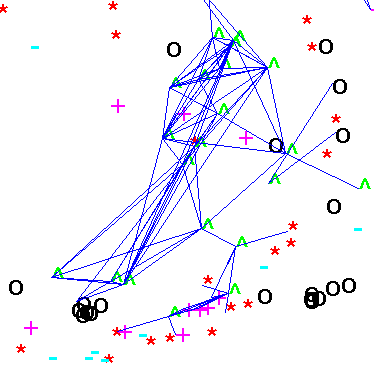}
    \label{fig:Group_diving_iter2crop}}
    \hfil
        \caption{The lines indicate the grouping of the class \emph{Diving} from the UCF11 dataset before and after pull groups together}
        \label{fig:GroupDiving}
\end{figure}
 Figure~\ref{fig:ExampleGroupTP} shows examples of the correct and incorrect classification of videos within the two circles of Figure~\ref{fig:InitalGroup_divingcrop} which is the relevant subsection of Figure~\ref{fig:InitalGroup_diving_Iter1}.
\begin{figure}[htbp]
    \centering
    \subfigure[]{
    \includegraphics[width = 0.29 \columnwidth]{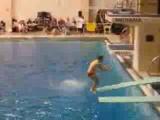}
        
    \label{fig:ExampleGroupTP_a}}
    \hfil
    \subfigure[]{
    \includegraphics[width = 0.29 \columnwidth]{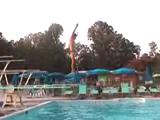}
        
    \label{fig:ExampleGroupTP_b}}
        \hfil
    \subfigure[]{
    \includegraphics[width = 0.29 \columnwidth]{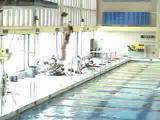}
        
    \label{fig:ExampleGroupTP_c}}
    \hfil
    \caption{Correct examples of image signatures from the \emph{Diving} class }
    \label{fig:ExampleGroupTP}
\end{figure}
\begin{figure}[htbp]
    \centering
    \subfigure[]{
    \includegraphics[width = 0.29 \columnwidth]{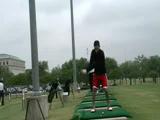}
        \label{fig:ExampleGroupTF_a}}
    \hfil
    \subfigure[]{
    \includegraphics[width = 0.29 \columnwidth]{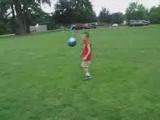}
    \label{fig:ExampleGroupTF_b}}        
    \hfil
    \caption{Incorrect examples of image signatures from the \emph{Diving} class}
    \label{fig:ExampleGroupTF}
\end{figure}
In this example, the incorrect examples generally contain the same vertical motion of diving as is the case of the golf swing~\ref{fig:ExampleGroupTF_a}, or the ball bouncing in Figure~\ref{fig:ExampleGroupTF_b} and therefore are incorrectly grouped and classified as \emph{diving} also. 

\subsection*{Pulling the groups together}
Figure~\ref{fig:InitalGroup_divingcrop} shows two circled groups of the diving class, naturally grouped. However, they contain incorrect examples and form two separate groups. The user would like to ``pull" the two groups together. To achieve this, the user can select a subset of correct classifications from within the two circles, and also 1 or 2 incorrect groupings. The mining will identify common elements of the true image signatures against the negative subset, and accentuate those elements in all the image signatures in the dataset. This will pull the true image signatures closer while at the same time ungrouping the negatively grouped image signatures. Figure~\ref{fig:Group_diving_iter2crop} shows the groupings after selecting six videos within the two marked circled groups, the grouping within the true examples of the class has increased and is reflected in the increased accuracy of correctly grouping \emph{diving} examples by 10\%. Also, some the incorrect links were removed as the correct links have increased in strength.

\subsection*{Pushing apart Groups}

\begin{figure}[htp]
    \centering
    \subfigure[Initial greedy clustering result of the class Jumping]{
    \includegraphics[width = 0.45 \columnwidth]{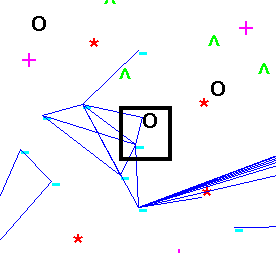}        
    \label{fig:Group_jumping_iter1}}
    \hfil
    \subfigure[Grouping of jumping class after pulling together two separate groups]{
    \includegraphics[width = 0.45 \columnwidth]{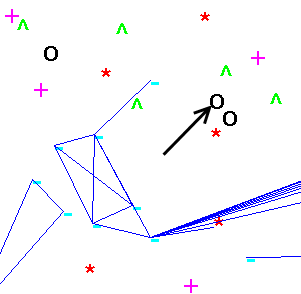}
    \label{fig:Group_jumping_iter2}}
    \hfil
        \caption{The lines indicate the grouping of the class; Jumping from the UCF11 dataset before and after pull groups together}
        \label{fig:GroupJumping}
\end{figure}

The approach can also be used to push apart incorrectly grouped videos. Within the box in Figure \ref{fig:Group_jumping_iter1} the circle and the horizontal line classes of videos are incorrectly classified as the same group. Therefore, the image signatures from these two videos are selected and mined to identify elements that occur in both image signatures. The identified elements are removed from all image signatures in the dataset, which reduces confusion between the two videos from the different classes and therefore the the set overlap of these image signatures which causes them to move apart visually, (as shown by Figure~\ref{fig:Group_jumping_iter2}).
The pushing apart of the incorrectly grouped image signatures in Figure~\ref{fig:GroupJumping} reduces the confusion rate of the jumping class by 5\%. The iterative process of pushing apart and pulling together of the image signatures continues, and this increases the overall accuracy of the correctly grouped media on the UCF11 dataset from a baseline figure of 60.4\% to 81.7\%, in only 15 iterations. Also, it should be noted, that the actual feature classifiers making up the image signatures have been learnt on the KTH training dataset. This serves to highlight that features learnt for classification, may not be the best features for grouping or clustering using simple distance metrics, but through the use of signatures and online learning, these features can be reweighed appropriately to achieve state-of-the-art performance. 

\subsection*{Comparison to other approaches}

A 6 fold cross validation is applied to the dataset to allow comparison to other traditional approaches applied to the UCF11 dataset. The training subsets were used to adjust the image signatures by performing 15 iterations, with up to 90 labelled training videos, with 5 correct and an incorrect classification selected ($<$5T,1N$>$), and the complete process is repeated 20 times, with the mean taken. The test subset was classified using the nearest neighbour assignment to the closest class. Table~\ref{tab:YoutubeAcc} shows the average results for our signature min-Hash approach compared to other recently published results on the same dataset.

\begin{table}[htbp]
\caption{Accuracy on UCF11 dataset}
  \begin{center}
\begin{tabular}{|c||c|c|c|c|}
  \hline
    Approach                                            &  Iter             &Accuracy       & Train GT     &    $\sigma$    \\
                                    &            &               &  Vids Used     &                        \\ \hline
    Bregonzio~\cite{BregonzioBMVC10}&      -     &63.1\%            &   1122              &        -                \\ 
    Liu~\cite{LiuCVPR09}               &      -     &71.2\%            &   1122              &        -                \\     
    Cinbis~\cite{IkizlerECCV10}     &      -     & 75.2\%        &   1122              &        -                \\    \hline \hline
    Baseline -Hash                &            0      &    56.4\%         &    0                 &        3.1\%        \\ \hline
   Sig min-Hash                   &        5   &   61.4\%      & 30           &        24.6\%    \\ 
   Sig min-Hash                   &        10  &   84.5\%      & 60           &        13.9\%  \\ 
   Sig min-Hash                   &        15  &\textbf{86.7\%}& 90           &        4.3\%        \\                            
   Sig min-Hash                   &        20  &\textbf{86.7\%}& 120          &        0.2\%      \\ \hline     
\end{tabular}
\end{center}
\label{tab:YoutubeAcc}
\end{table}

\subsubsection{Hollywood2 Dataset}

The final video dataset examined is the  Hollywood2 dataset~\cite{MarszalekCVPR09}. It consists of 12 action classes; \emph{AnswerPhone, DriveCar, Eat, FightPerson, GetOutCar, HandShake, HugPerson, Kiss, Run, SitDown, SitUp, StandUp} with around 600,000 frames or 7 hours of video sequences split evenly between training and test datasets.
 The image signatures for this dataset are based on dense
trajectory features ~\cite{wang2013denseTraj}, an optical flow based feature descriptor consisting of Trajectory, HOG, HOF and MBH. 
The dimension of the descriptors is 30 for Trajectory,
96 for HOG, 108 for HOF and 192 for MBH, giving a base feature size of 426. We then train a 4000 element
codebook using 100,000 randomly sampled feature
descriptors with k-means, which when converted to the image signature, contain around 5100 elements on average per video sequence. The clean train and
test partitions proposed by Marszalek~\cite{MarszalekCVPR09} were used,
where there is a total of 810 specified videos within
the training subset spread over the 12 action classes.
In total 25 iterations of our approach was performed,
selecting five correct classifications and a single
incorrect classification at each iteration ($<$5T,1N$>$),
moreover, this process was repeated for 20 user runs and
averaged. The adjusted image signatures were then applied to the 884 test sequences and classified using the nearest neighbour assignment. To fully
compare the online/active learning method with a
traditional train/test approach, an additional test was
performed where the dense feature trajectories, and
full standard labelled training data was used with
the initial image signatures and trained through the
APriori data mining, to form a separate classifier for
each class. This approach is indicated as DM DenseTraj
in Table~\ref{tab:Hollywood2Acc}. Table~\ref{tab:Hollywood2Acc} also shows the accuracy for the
baseline and each iteration of learning in comparison
to other state-of-the-art approaches.

\begin{table}[htbp]
\caption{Accuracy of the Hollywood2 dataset}
   \begin{center}
\begin{tabular}{|c|c|c|c|c|}
   \hline
     Approach                           & Iter          & Acc     &Train GT         & $\sigma$    \\
                                                                                &                     &              &  Vids Used &        -                \\ \hline
     Marszalek~\cite{MarszalekCVPR09}        &     -     & 35.5\%          &   810              &        -                \\
     Han~\cite{HanICCV09}               &     -     & 42.1\%          &      810              &        -                \\
     Wang~\cite{WangBMVC09}             &     -     & 47.7\%          &   810              &        -                \\
     Gilbert~\cite{GilbertICCV09}       &         - &50.9\%        &   810            &        -                \\
     Vig~\cite{VigECCV12}               &        -  & 59.4\%       &        810   &        -              \\
     Jain~\cite{JainCVPR13}             & -              &    62.5\%       &   810                 &        -                \\   
     Wang~\cite{WangICCV2013_AprioriActrec}& -      &    64.3\%       &        810     &        -                \\  
     Gilbert~\cite{GilbertACCV14}		&	-		&	64.5\%		& 810			& -	\\
     Lan~\cite{Lan_2015_CVPR}           &      -    &\textbf{68.0\%}&   810 &  - \\ \hline \hline
     DM DenseTraj                & -              &65.1\%           &        810     &        -                \\
     Baseline -Hash                  & 0              &26.9\%           &    0                &        4.5\%        \\ \hline
     Sig min-Hash                       &     5          &39.0\%           &   30             &        21.3\%    \\
     Sig min-Hash                       &     10         &45.2\%           &   60             &        16.9\%    \\
     Sig min-Hash                       &     15         &57.4\%           &   90             &        6.3\%        \\
     Sig min-Hash                       &     18         &64.9\%           &   108             &        1.2\%        \\
     Sig min-Hash                       &     20         &64.9\%           &   120             &        0.4\%        \\
     Sig min-Hash                       &     25         &64.9\%           &   150                &        0.03\%    \\ \hline

\end{tabular}
\end{center}
\label{tab:Hollywood2Acc}
\end{table}
The final stable accuracy of 64.9\% is over double the original baseline of 26\%, using only 108 labelled videos, this compares favourably to the standard training approach using all 810 training videos, with a minimal difference in performance. Similarly, there is an increase in performance compared to other state of the art approaches such as Wang~\cite{WangICCV2013_AprioriActrec} and Vig~\cite{VigECCV12}. The performance increase over the approach by Wang which uses the same feature descriptors is due in part to the targeted training of the image signatures that is possible by our method. We can focus on the areas of confusion to increase the performance, coupled with the efficient exhaustive training methodology of the APriori data mining. Furthermore, it can be seen that the classification performance is stable after the 18th iteration, ensuring that the image signatures are not over fitted to the training data.

\subsubsection{Active Learning Datasets}
To provide a comparison of our method
against a standard active learning approach, we use a
dataset from the UCI repository Iris~\cite{Lichman2013}. Iris contains
samples from the species of three flowers, the numerical descriptor is based on length and width criteria
of their blossoms. There are 150 samples and ten fold
cross-validation was performed from 10 user trials of
15 iterations, where for each iteration a ratio of ($<$4T,1N$>$) examples were
selected (see table ~\ref{tab:IrisResults}).
\begin{table}[htbp]
\caption{Accuracy of the Iris dataset}
   \begin{center}
\begin{tabular}{|c|c|c|c|c|}
   \hline
     Approach                           & Iter          & Acc     &Train GT         & $\sigma$    \\
                                                                                &                     &              &  Imgs Used &                        \\ \hline
     Lughofer~\cite{lughofer2012hybrid}    &     -     & 82.3              &   15            &        19.77        \\
     Lughofer~\cite{lughofer2012hybrid} &     -     & 89.51          &      30           &        14.33        \\
     Lughofer~\cite{lughofer2012hybrid} &     -     & 90.78\%         &   45           &        11.49     \\
     Lughofer~\cite{lughofer2012hybrid} &         - & 92.95\%      &   75         &        11.72        \\
     Lughofer~\cite{lughofer2012hybrid} &        -  & 94.0\%       &   150          &        7.34      \\ \hline \hline
         Baseline -Hash                                    &        0  & 56.7\%       &   0            &        47.1      \\
         Sig min-Hash                                             &        6  & 89.5\%       &   30            &        10.2      \\
         Sig min-Hash                                             &        9     &\textbf{95.8\%}&   45            &        2.8        \\
         Sig min-Hash                                         &        15 & \textbf{95.8\% }&   75            &        0.01      \\ \hline

\end{tabular}
\end{center}
\label{tab:IrisResults}
\end{table}

While the dataset is simple, it still allows compassion with an active learning approach~\cite{lughofer2012hybrid}, comparing performance with increasing amounts of labelled data, as the number of iterations increases. Also, our approach shows the reduced amount of labels required and reduced $\sigma$  to provide state of the art performance on this dataset.

\subsection{Cluster Purity}
%

\begin{figure}[htp]
    \centering
    \subfigure[]{
    \includegraphics[width = 0.45 \columnwidth]{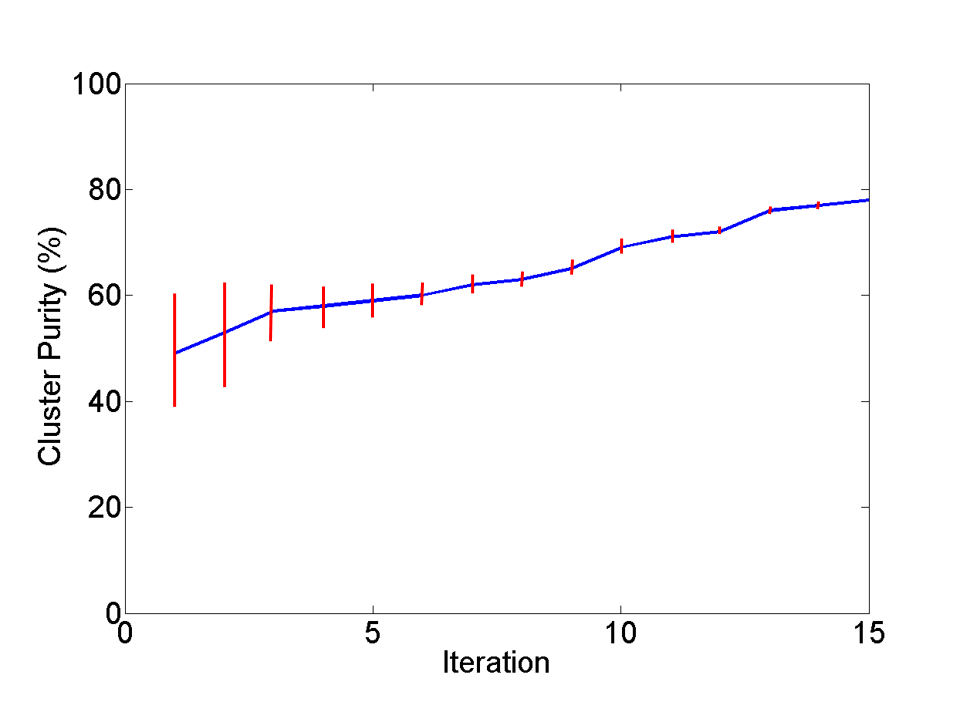}
    \label{fig:youtubeClusterPurity}}
    \subfigure[]{
    \includegraphics[width = 0.45 \columnwidth]{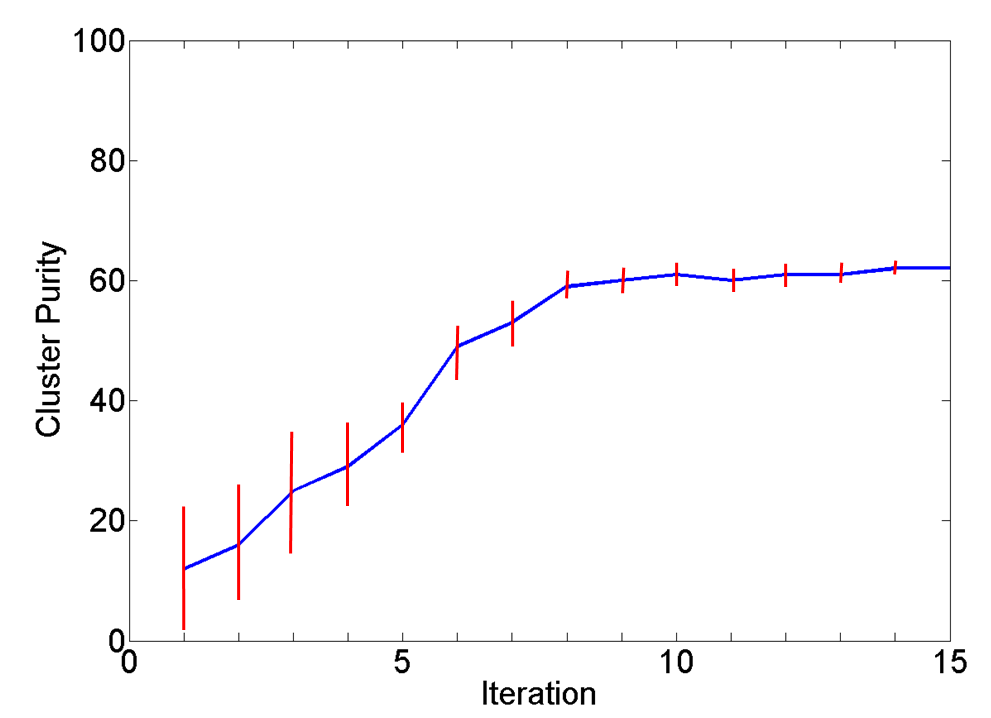}
    \label{fig:HollywoodClusterPurity}}
    \caption{(a)Cluster purity on UCF11 over 15 iterations, for 20 runs, (b) The Hollywood2 cluster purity and error bars over 15 iterations, for 20 unique runs }
    \label{fig:ClusterPurity}
\end{figure}

Figure~\ref{fig:youtubeClusterPurity} shows the cluster purity of the UCF11 dataset over 15 iterations, for 20 runs, using $<$5T,1N$>$ selections of the training data as employed in the results above


A similar figure is also shown for the Hollywood2 dataset's cluster purity, in Figure~\ref{fig:HollywoodClusterPurity} together with error bars. The error bars indicate the standard deviation of 20 runs of grouping the Hollywood2 dataset, using $<$5T,1N$>$ selections of the training data as utilized in the results offset.


Both of these video datasets initially have low cluster purity especially in the case of the Hollywood2 dataset, illustrating the complexity of the dataset. However, as the iterative process is carried out, the purity rapidly increases. This indicates that not only is the approach able to achieve a high accuracy in comparison to other approaches, it can also produce relatively pure groups of media, with little cross contamination. The error bars initially are quite large, with a standard deviation of around 10\%, however after around eight iterations this is reduced to 1\% or 2\%. This shows that the examples the user selects can have a considerable effect on similarity initially. Also, the random process of the min-Hash will affect the variability, but as the number of examples increases this variability decreases. Further iterations show no further progress but also no over fitting or decrease in cluster purity. No further increase or change in performance is due to no new co-occurring mined rules being identified. Therefore, no changes are made to the image signatures.

\subsection{Computational costs}

The min-Hash algorithm is designed to be invariant to the length or complexity of the image signatures and is dependant upon the quantity of image signatures or size of the dataset. Similar characteristics are present with the APriori data mining, designed for large sets of transactions. The use of these data mining tools allows for real-time operation on some of the smaller datasets. Table~\ref{tab:Speed} shows the average computation time for an iteration for each dataset. There is also a user ``thinking time'' time, required to select each subset group, however, due to the MDS visualisation, this is less than 10 seconds per iteration.
 
\begin{table}[htbp]
\caption{Computational Time of datasets}
  \begin{center}
\begin{tabular}{|c||c|c|c|}
  \hline
    Dataset                                            &  Dataset Size      & Img Sig Size & Iter Time   \\ \hline
    15 Scene                                           &            4486    &    512     &        31        sec                          \\
    Caltech101                                         &     5050           &        2150        &        75 sec                                    \\    
    ImageTag                                           &       800          &        9250      &        50 sec                                \\
    FG-NET                                                                                         &        1001                        &     1450        & 17 sec   \\
        UCF11                                                &        1200                         &        3108        &        63 sec                                    \\
    KTH                                                        &    768                         &   1204            &      25 sec                                    \\
    Hollywood2                                      &             884                         &        5100        &        48 sec                                    \\
    Iris                                                    &            150                         &        20          &   3 sec                                        \\ \hline
\end{tabular}
\end{center}
\label{tab:Speed}
\end{table}

\section{Conclusions}
We have presented a unique approach that intelligently
employs user input to identify the areas of
confusion within large datasets, allowing learning to
iteratively refine distances between different media
types. The use of the min-Hash, APriori, and image
signature containers, allow the approach to operate
accurately and efficiently despite size, type or representation.
This is illustrated by the approach being
able to process, cluster, group and visualise the entire
UCF11 dataset of over 1200 videos in just over 1
minute. To further improve the performance of the
approach it would be possible to fuse other high and
low level feature types into the image signature to
capture additional information that the dataset image
and videos contain. This type of performance increase
was shown by the addition of the text feature to the
image feature for the ImageTag dataset. A future extension of the work would be to intelligently influence the user's selection process in the iterations by automatically identifying "probable" areas of confusion in the data and highlighting these to the user.

\subsection*{Acknowledgements}
This work is supported by the EPSRC projects Making Sense (EP/H023135/1) and Learning to Recognise Dynamic Visual Content from Broadcast Footage (EP/I011811/1).

\bibliographystyle{elsarticle-num}  
\bibliography{Bibliography}

\end{document}